\documentclass[runningheads]{llncs}
\usepackage{amsmath,amssymb,amsfonts,textcomp}

\usepackage{fancyvrb}
\usepackage[dvipsnames]{xcolor}

\usepackage[T1]{fontenc}
\usepackage[ascii]{inputenc}
\usepackage{array}
\usepackage{url}
\usepackage{cite}
\usepackage[pdftex]{graphicx}  
\usepackage{footnote}
\usepackage{hyperref}
\usepackage{algpseudocode}
\usepackage{algorithm}
\usepackage{float}
\usepackage{listings}
\usepackage{ifthen}
\usepackage[italic,defaultmathsizes,noendash]{mathastext}
\MTnonlettersobeymathxx
\MTeasynonlettersobeymathxx
\usepackage{tabularx,ragged2e}

\newcommand{\mycomment}[1]{}
\newcommand\ifnotempty[2]{\ifthenelse{\equal{\unexpanded{#1}}{}}{}{#2}}
\newcommand\tsub[1]{\ensuremath{{}_{\text{#1}}}}
\newcommand\calsubscript[1]{\ensuremath{{}_{\cal{#1}}}}

\makeatletter
\newcommand\arraybslash{\let\\\@arraycr}
\makeatother
\newcommand{\openps}{\texttt{(}}
\newcommand{\closeps}{\texttt{)}}
\newcommand{\hash}{\,\#\,}

\makeatletter
\newcommand{\verbatimfont}[1]{\renewcommand{\verbatim@font}{\ttfamily#1}}
\makeatother

\newcommand\sldots{\,\ldots\,}

\newcommand\angles[1]{\ensuremath{\langle{#1}\rangle}}

\newcommand\tup{t\tsub{1}\ldots t\tsub{n}}

\newcommand\ttarrow{\textup{\texttt{->}}}
\newcommand\ttarrowdep{\textup{\texttt{+>}}}
\newcommand\ttimp{\,\textup{\texttt{:\!-}}\ }

\newcommand\textbfit[1]{\textbf{\textit{#1}}}

\newcommand\funcbfit[4][]{\ensuremath{\textbfit{#2}_{\text{#3}}^{#1}\ifnotempty{#4}{\left(\Mathtt{#4}\right)}}}
\newcommand\funcsf[4][]{\ensuremath{{\Mathsf{#2}}_{#3}^{#1}\ifnotempty{#4}{\left({#4}\right)}}}
\newcommand\functt[2]{\ensuremath{\Mathtt{{#1}}\ifnotempty{#2}{\Mathtt{(}{#2}\Mathtt{)}}}}

\newcommand\tvalfunc[4]{\mbox{\ensuremath{\textit{TVal}_{#1}^{#4}\ifnotempty{#2}{\left(#2\right)}\ifnotempty{#3}{=\textbf{#3}}}}}
\newcommand\tvaltt[3][\ip{}]{\tvalfunc{#1}{\Mathtt{#2}}{#3}{}}

\newcommand\ipmap[2][]{\funcbfit{I}{#1}{#2}}

\newcommand\ipsoa[1]{\ensuremath{\ipmap[psoa]{#1}}}
\newcommand\dom[2][]{\funcbfit{D{#2}}{#1}{}}

\newcommand\dind[1][]{\dom[ind]{#1}}

\newcommand\true{\textbf{t}}

\newcommand\mapprl[1]{\funcsf[]{\rho}{}{#1}}

\mycomment{
\tikzset{
    every node/.style={font=\footnotesize},
    oidnode/.style={orange},
    cnode/.style={red},
    clsnode/.style={blue},
    snode/.style={magenta,font=\tiny},
    redge/.style={semithick,-stealth},
    bedge/.style={* new-stealth,semithick,arrow head=1mm}
}

\tikzset{
    every node/.style={font=\footnotesize},
    oidnode/.style={orange},
    cnode/.style={red},
    clsnode/.style={blue},
    snode/.style={magenta,font=\tiny},
    blanknode/.style={white},
    redge/.style={semithick,-stealth},
    bedge/.style={* new-stealth,semithick,arrow head=1mm},
    sedge/.style={o new-stealth,semithick,arrow head=1mm}
}
}

\title{Perspectival Knowledge in PSOA RuleML: \\
Representation, Model Theory, and Translation}

\titlerunning{Perspectival Knowledge in PSOA RuleML}

\date{}

\begin{document}

\author{Harold Boley, Gen Zou}

\institute{Faculty of Computer Science, University of New Brunswick, Fredericton, Canada\\
\email{\{harold[DT]boley, gen[DT]zou\}[AT]unb[DT]ca}
}

\maketitle

% ...

\begin{abstract}
In Positional-Slotted Object-Applicative (PSOA) RuleML, a predicate application
(atom) can have an Object IDentifier (OID) and descriptors that may be positional
arguments (tuples) or attribute-value pairs (slots). PSOA RuleML explicitly specifies for each descriptor
whether it is to
be interpreted under the perspective of the predicate in whose scope it occurs.
This predicate-dependency dimension refines the design space
between oidless, positional atoms
(relation\-ships) and oidful, slotted atoms (framepoints): While relationships
use only a predicate-scope-sensitive (predicate-dependent) tuple and
framepoints use only predicate-scope-insensitive (predicate-independent) slots, PSOA
uses a systematics of orthogonal constructs also permitting atoms with
(predicate-)independent tuples and atoms with (predicate-)dependent slots. This
supports advanced data and knowledge representation where, e.g., a slot attribute
can have different values depending on the predicate.
PSOA
thus extends classical object-oriented
multi-membership and multiple inheritance.
Based on objectification, PSOA laws are explicated: Besides unscoping and centralization,
the semantic restriction and implemented transformation of describution
permits the rescoping of one atom's independent descriptors to another atom with
the same OID but a different predicate.
For inheritance, default descriptors are realized by rules.
On top of a basic metamodel and a new Grailog visualization,
PSOA's use of the atom systematics for facts, queries, and rules is explained.
The presentation and (XML-)serialization syntaxes of PSOA RuleML are introduced.
Its model-theoretic semantics is formalized by extending the interpretation
functions to accommodate dependent
descriptors. The open-source PSOATransRun system
realizes PSOA RuleML
by a translator to runtime predicates, including for dependent tuples (prdtupterm) and slots (prdsloterm).
Our tests show efficiency advantages of dependent and tupled modeling.
\end{abstract}

\section{Introduction}\label{intro}


\mycomment{
In the field of Artificial Intelligence (AI), context has recently become a central R \& D topic vis-\`a-vis increasing data and knowledge volumes: AI applications should benefit from
contextual representation, reasoning, and learning.

John McCarthy proposed formalizing the notion of context by adding a context
parameter to the functions and predicates in each axiom, where a $\leq$ relation between context names allows inheritance of facts [McCarthy, 1987].
This proposal led to various contextual logics.

In object-centered logics such as F-logic, RIF-BLD, and PSOA RuleML, the Object IDentifier (OID) can be used as a context parameter in two ways.

(1) A {\it propositional OID} can be added to a predicate application.
For example, an atom like ``The department of John is Math'' may be represented as department(John Math); Then ``C1: The department of John is Math'', with propositional OID C1, becomes C1\#department(John Math).

(2) One of the predicate application's arguments can be extracted as an {\it individual OID} and a predicate specify the context.
For example, an atom like ``The department of John is Math'' may also be represented, with the individual OID John, as John\#Top(dept->Math); Then ``The department of John as a Student is Math'', with contextual predicate Student, becomes John\#Student(dept+>Math).
}

\begin{figure*}
\centering
\includegraphics[width=12.5cm]{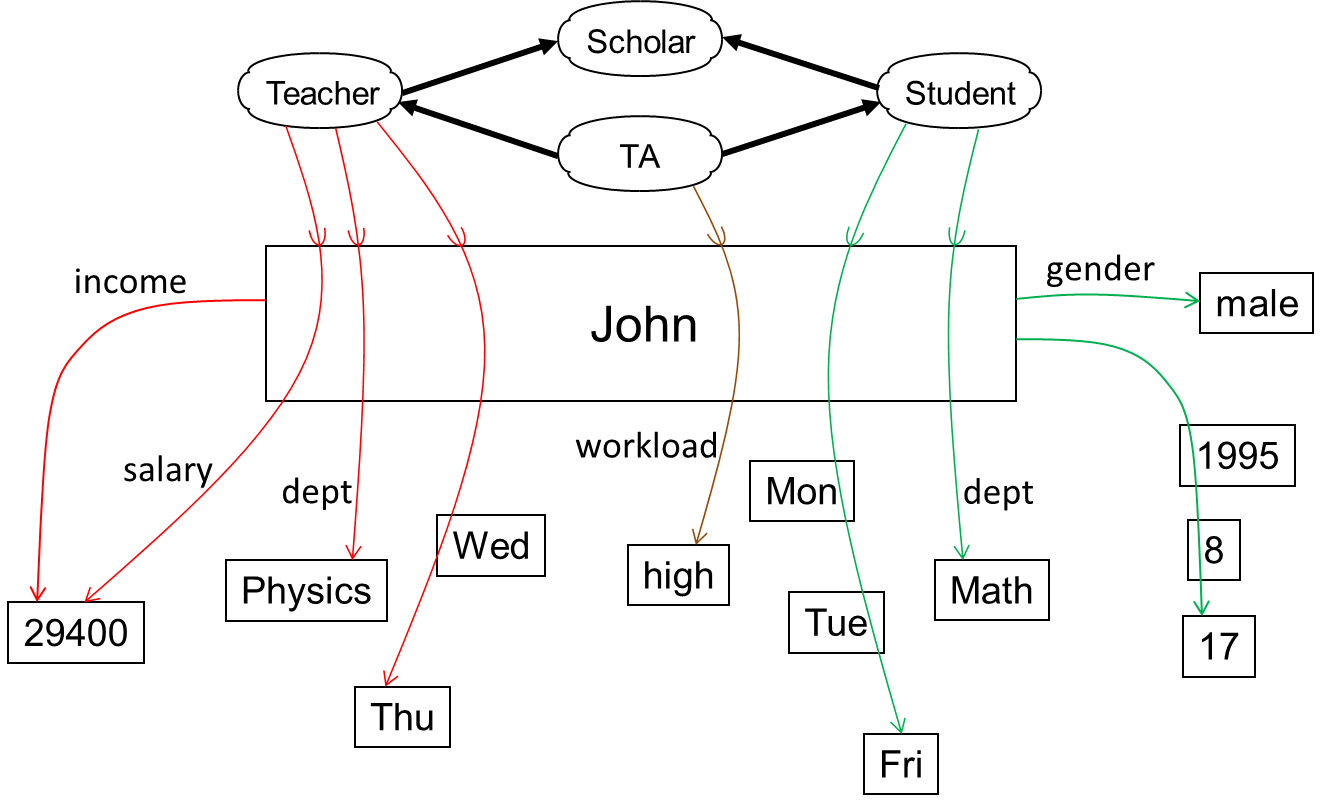}
\vspace{-0.8cm}
\caption{Rich TA example of independent/dependent facts in Grailog: OID John 
described independent of / dependent on predicates Teacher, TA, Student.}
\label{TAexample}
\end{figure*}

In advanced Artificial Intelligence (AI) Knowledge Bases (KBs),
the related \linebreak
notions of ``context'' and
``perspective'' are both called for. While a {\it context} mechanism~\cite{Gratton13} allows to partition the clauses of a KB, {\it perspective}, as introduced here, allows to describe the same Object IDentifier (OID) differently with multiple clause conclusions
-- e.g., predicate applications (atoms) used as facts -- having different predicates
(cf. Fig.~\ref{TAexample}'s OID {\tt John} with predicates {\tt Teacher},$\;${\tt TA},$\;${\tt Student}).

A form of {\it contextualized} KBs has been available in Positional-Slotted Object-Applicative RuleML (PSOA RuleML)~\cite{Boley11d,Boley15b,ZouBoley15,ZouBoley16,Zou18,Boley18e}\footnote{\url{http://wiki.ruleml.org/index.php/PSOA_RuleML}} as realized by PSOATransRun since Version 1.2, allowing (1) constants that are local to each KB and (2) a merging {\tt Import} statement that will rename apart  
local constants from multiple KBs.
Reciprocally, the current paper focuses on the topics of representation, model theory, and translation for {\it perspectival} KBs (fact \& rule clauses) and queries as explicitly used since PSOA RuleML 1.0 and realized since PSOATransRun 1.3.

Our notion of perspective is part of
a novel orthogonal categorization (henceforth: systematics) for \textbf{p}ositional-\textbf{s}lotted \textbf{o}bject-\textbf{a}pplicative (\textbf{psoa})\footnote{We use the upper-cased ``PSOA'' as a qualifier for the language and the lower-cased ``psoa'' for its terms, i.e. atoms or expressions, and parts of its terms, e.g. descriptors.} atoms, which constitutes the basic PSOA RuleML metamodel of Fig.~\ref{Metamodel} in Appendix~\ref{sec:psoametamodel}.
Besides their use as data facts and -- often with variables -- as queries, psoa atoms occur in rule conclusions and conditions (because of the wide use of these formulas, when the intent is obvious we will frequently shorten ``psoa atom'' to ``atom'').

PSOA RuleML allows an {\it atom}
-- with an optional OID --
to apply a predicate
--
which, as a class, types that OID
-- 
to a bag (multiset) of tupled descriptors, each representing an argument 
sequence, and 
to a bag of slotted descriptors, each representing an attribute-value pair. 
Extending these dimensions for OIDs and descriptor varieties by a dimension for descriptor (predicate-)dependencies, PSOA RuleML atoms will be visualized, in Fig.~\ref{TAexample}, and explained with this oidful example (one shared OID is shown as a large box) before giving  symbolic forms.

We introduce ``Rich TA'' as a running example of taxonomy-augmented data. Fig.~\ref{TAexample} enriches the classical Teaching Assistant (TA) example for multiple inheritance~\cite{Taivalsaari95} and multi-membership
in object-oriented programming languages and databases. Previewing chief aspects, our enriched AI-KB version will illustrate where PSOA's expressivity increases compared to classical related work~\cite{Taivalsaari95,Pernici90}:
\vspace{-0.4em}
\setlength{\baselineskip}{1.15em}
\begin{itemize}
\item Permission of perspectives without diminishing the uniform notion of ``class'', as by an additional notion of (an individual playing a) ``role'' in the sense of \cite{Pernici90}\footnote{To be distinguished from the notion of ``role'' in the sense of ``property'' as used in
Description Logic (standardized as OWL 2: \url{http://www.w3.org/TR/owl-overview}),
``object-holes'' in Object-Role Modeling (\url{http://www.orm.net/pdf/ormwhitepaper.pdf}),
and ``association ends'' in the Object Constraint Language (\url{https://st.inf.tu-dresden.de/files/general/OCLByExampleLecture.pdf}).
},
which would hinder uniform knowledge representation for, e.g., sorted logic, Description Logic~\cite{DoniniLenzerini+98}, F-logic~\cite{flogic95}, N3Logic~\cite{Berners-LeeConnolly}, as well as ConceptBase's~\cite{JarkeGallersdoerfer+95} and PSOA's logics,
also suffering from the lack of a clear ``class''/``role'' boundary when descending a taxonomy like 
{\tt Top}, ..., {\tt Person} (``class''), {\tt Scholar} (``class'' that could be a ``role'' with a sibling 
{\tt Vegetarian} ``role''), {\tt Teacher} or {\tt Student} (sibling ``roles'' in \cite{Pernici90}), and {\tt TA}.
\item Distinction of predicate-independent and predicate-dependent descriptors, leading (when made \textcolor{violet}{for all} descriptors) to predicate-independent/dependent and (when made \textcolor{olive}{existentially}) to non-/perspectival atoms, clauses, and KBs,
so that the same OID, here {\tt John}, via certain atoms -- e.g. used as facts -- 
can be described under no perspective (equivalently, under the vacuous {\tt Top}, i.e. {\it root}, perspective) and via other atoms under
(non-{\tt Top}) perspectives, here {\tt Teacher}, {\tt TA},
{\tt Student}, with perspectives realized by predicates (``classes'').
\item Clausal extensibility of factual data by rule ({\it conclusion} {\tt :-} {\it condition}) knowledge --
possibly, as in Section~\ref{sec:presentationsyntax}, Fig.~\ref{SampleKB}, centered on an OID variable --
for inferencing,
such as to integrity-check existing data or to derive new data from it
(e.g., rather than storing a {\tt workload} fact for {\tt John}, as in Fig.~\ref{TAexample}: \linebreak
deriving it, again perspectivally for any {\tt TA}, based on other facts, as in Fig.~\ref{SampleKB}).
\end{itemize}

Constituting a portion of what can be regarded as an individual's ({\tt John}'s) \linebreak
``Perspectival Knowledge Graph'', Fig.~\ref{TAexample} generalizes earlier Grailog~\cite{Boley13} visualizations of PSOA KBs~\cite{Boley15b}\footnote{\url{http://wiki.ruleml.org/index.php/Grailog\#Family_Example}}
for accommodating the dependency dimension.

In the upper part, it shows a diamond-shaped taxonomy of four oval-like predicates\footnote{PSOA's taxonomies represent predicate subsumptions much like class subsumptions, where the root predicate {\tt Top} is always understood to subsume all other predicates. Thus, subpredicate arcs and facts linking from non-{\tt Top} subtaxonomy roots to {\tt Top} are not normally shown in, respectively, taxonomy DAGs such as Fig.~\ref{TAexample} (with subtaxonomy root {\tt Scholar}) and their symbolic forms such as {\tt (KB1)}-{\tt (KB3)} of Section~\ref{factsandqueries}.} -- {\tt Scholar}, {\tt Teacher}, {\tt Student}, and {\tt TA} -- connected by heavy arcs understood to be implicitly labeled with {\tt subpredicate}, where {\tt TA} --
connecting to both {\tt Teacher} and {\tt Student} -- exemplifies multiple inheritance.

In the lower part,
featuring directed-hypergraph-visualized data,
three of these predicates --
all except {\tt Scholar} -- are used to spawn dependent descriptors for perspectival representation.
For this, it uses hyperarcs starting with a predicate labelnode, e.g. {\tt Teacher}, pointing to -- with an element-symbol arrow head -- \linebreak
and cutting through an optional OID node, e.g. {\tt John}, and cutting through any further nodes in sequence before pointing to the last node, with all nodes being rectangular. Optional labels on these descriptor hyperarcs, e.g. {\tt dept}, are slot names, thus distinguishing slot hyperarcs from tuple hyperarcs.
E.g., the {\tt Teacher} hyperarcs indicate, from right to left, that -- under the perspective of being a {\tt Teacher} -- {\tt John}
is associated with (a length-2 tuple, in standard chronological order, for) {\tt Wed} followed by {\tt Thu},
is in the {\tt dep}(artmen){\tt t} of {\tt Physics}, and
has a {\tt salary} of {\tt 29400}. 
On the far left, a labeled arc, starting directly at the OID, records {\tt John}'s (total) {\tt income} (also) as {\tt 29400} -- independently of, e.g., the {\tt Teacher}, {\tt TA}, and {\tt Student} perspectives.

\setlength{\baselineskip}{1.15em}
Two complementary {\bf methods of creating atoms} from these descriptor (hyper)arcs exist, besides various other groupings:
\textbf{(i)} for {\it single-descriptor atoms}, each made of one (hyper)arc; \textbf{(ii)} for {\it dependence-concentrated atoms}, each made of all the hyperarcs starting with a common predicate and continuing with a common OID, as well as of
zero or more
(hyper)arcs
starting only at this OID. \linebreak
Using the descriptor (hyper)arcs
discussed
so far: the unique, atom-size-mini\-mizing method (i) creates four atoms, where the red color is immaterial; \linebreak
the non-unique, atom-count-minimizing method (ii) 
creates one atom, where the red color serves for large-atom chunking of these descriptors, chosen to include the predicate-independent {\tt income} slot (but no other independent descriptors).\footnote{The single-descriptor atoms according to method (i) can be obtained from arbitrary \linebreak
atoms by the describution {\bf law} and {\bf transformation} of Sections~\ref{unscopingtodescribution} and \ref{sec:implement}, resp., based on a 
semantic {\bf restriction} in Section~\ref{sec:semantics};
the zero-or-more-descriptor atoms according to method (ii) correspond to those obtained by
centralization in Section~\ref{unscopingtodescribution}.}

\setlength{\baselineskip}{1.15em}
The remaining (hyper)arcs are similar except that in the --$\;$green-grouped$\;$-- 
atom -- under the perspective of the start labelnode {\tt Student} -- {\tt John}
is associated with, e.g., (a length-3 tuple for) {\tt Mon} followed by {\tt Tue} and {\tt Fri}, and that \linebreak
-- independently of predicates (thus applicable to {\tt John} in an `absolute' manner) -- \linebreak
he is associated with (a length-3 tuple for) {\tt 1995} followed by {\tt 8} and {\tt 17}.\footnote{A tuple (hyperarc) can be seen as a shortcut for a tuple-valued slot (hyperarc) having the {\tt Top}-predicate-complementing implicit `vacuous' name (label) {\tt prop}(erty), which \linebreak
could
be specialized
here
to slot names like {\tt dop} -- for the (dependent) days-of-presence of a scholar -- and {\tt dob} -- for the (independent) date-of-birth of a person.
A multi-tuple \linebreak
psoa term can expand its tuples to (in/dependent) multi-(tuple-)valued {\tt prop} slots.
\label{footnote:prop}}  \linebreak
Generally, method (i) ignores any -- here, three -- 
colors while method (ii) uses them to indicate grouping of descriptors into atoms.

\setlength{\baselineskip}{1.15em}
Since {\tt John} is represented as an OID node pointed to and cut through by hyperarcs starting with three different predicates -- {\tt Teacher}, {\tt TA}, and {\tt Student} -- \linebreak
he is involved under these different perspectives. The ``pointing to'' also entails a multi-membership of {\tt John} in three predicates, here acting as classes.
Abbreviating ``under the perspective of'' to ``as a'' or ``asa'', 
we can generally say that ``asa entails isa'', where the ``isa'' of classical 
Semantic Nets is often called ``is member of'' in Semantic Technologies.
Notice that for perspectival (data and) knowledge, multi-membership cannot be reduced to multiple inheritance with a newly introduced common subpredicate such as {\tt TA} underneath {\tt Teacher} and {\tt Student}:
The very notion of perspective requires that an individual such as {\tt John} stays member of the predicates under whose perspectives it is represented.

\setlength{\baselineskip}{1.09em}
This section introduced the novel dependency dimension as part of a systematics with other dimensions for atoms in PSOA RuleML, 
illustrated by a visualized three-perspective example.
The subsequent Section~\ref{foundknowrep} will continue with symbolic perspectival fact \& rule representation and reasoning (through querying) in the abridged syntax of PSOA RuleML.
This will be followed, in Section~\ref{lawsknowtrans}, by equivalence laws for PSOA knowledge, with a subsection on default descriptors realized via default rules.
The paper will then proceed, in Section~\ref{sec:syntax}, to the appropriately augmented unabridged presentation syntax and the serialization syntax of PSOA RuleML 1.03.
Next, in Section~\ref{sec:semantics}, it will revise the parts of the model-theoretic semantics that are key to incorporating in/dependent descriptors.
Sections \ref{sec:syntax} and \ref{sec:semantics} establish PSOA as a logic.
Then, in Section~\ref{sec:implement}, the paper will discuss the PSOATransRun implementation of in/dependent descriptors, translating PSOA RuleML knowledge bases and queries to TPTP (PSOA2TPTP) or \linebreak
Prolog (PSOA2Prolog);
test results will be shown.
Finally, Section~\ref{sec:conc} will give conclusions and indicate directions of future work.
The often-referenced
Appen\-dix~\ref{sec:psoametamodel} conveniently wraps the metamodel, applying it to the Rich TA example. \linebreak
The examples of this paper bridge between theory and practice: They have been tested in the PSOATransRun instantiation targeting XSB Prolog,
and readers are encouraged to try and change some of them,
starting with the README\footnote{\url{http://psoa.ruleml.org/transrun/1.4.2/local/}}.

\vspace{-0.5em}
\section{Foundations of PSOA Knowledge Representation}\label{foundknowrep}

\vspace{-0.1em}
\setlength{\baselineskip}{1.15em}
In this section we discuss the foundations of knowledge representation in PSOA RuleML, advancing a concrete syntax to formalize KBs according to the metamodel of Appendix~\ref{sec:psoametamodel}, illustrated by Fig.~\ref{TAexample} of Section~\ref{intro}. We further give positive and negative query examples that provide informal proof-theoretic \linebreak
semantics in preparation for the formal model-theoretic semantics in Section~\ref{sec:semantics}. 
We also discuss modeling approaches to reduce dependent to independent slots.

\setlength{\baselineskip}{1.15em}
$\!\!\!$An {\it (in)dependent atom/clause/KB} \textcolor{violet}{has only} (in)dependent descriptors/atoms/ 
clauses.
A {\it (non-)perspectival atom} does (not) \textcolor{olive}{have some} dependent descriptor; a {\it (non-)perspectival clause/KB} does (not) \textcolor{olive}{have some} perspectival atom/clause.

\vspace{-0.6em}
\subsection{Formal Facts and Their Querying}\label{factsandqueries}

\setlength{\baselineskip}{1.15em}
To formalize the notions of Section~\ref{intro}, we complement the {\it visualization syntax} used there by a {\it presentation syntax}, developing the one in \cite{Boley15b}. This abridges the
EBNF syntax of Section~\ref{sec:presentationsyntax}, employed
by
the
PSOATransRun
system, omitting the {\tt RuleML} and {\tt Assert} wrappers from KBs
as well as the optional ``{\tt \_}'' prefix from most local constants such as {\tt \_John}, except for
(objectification) integers.\footnote{\setlength{\baselineskip}{1.0em}For example,
any of the three Fig.~\ref{TAexample} hyperarcs starting with the predicate
{\tt Teacher} and pointing -- via an element (``$\in$'') arrow head --
to the OID {\tt John}
can be
symbolically represented as a {\it membership} of
{\tt John} in (indicated by ``\texttt{\#}'') {\tt Teacher}
by the oidful empty atom
{\tt John\#Teacher()},
often shortened to {\tt John\#Teacher},
e.g. as a fact, query, or in a (conclusion or condition of a) rule.
The corresponding oidless empty atom
{\tt Teacher()}
will be {\it objectified}~\cite{Boley11d,Boley15b,ZouBoley16} by PSOATransRun,
e.g. when used as a fact yielding
{\tt \_}$j${\tt \#Teacher()}, $j\geq1$,
where ``{\tt \_}$j$'' is generated as the fresh local positive-integer
OID
(employed as a Skolem constant, similarly to an RDF blank node) having the minimal $j$.
Objectification works the same for non-empty atoms, since it does not involve their (dependent or independent) descriptors.
The ``{\tt \_}'' of ``{\tt \_}$j$'' can never be omitted.
For any integer $j$, {\tt \_}$j$ $\neq$ $j$.
No other numbers are used with a ``{\tt \_}'' prefix.
}
\mycomment{
For example, the sentence ``It is sunny.'' in PSOA can be asserted and queried using the oidless empty atom {\tt Sunny()}.
Likewise, ``In situation 1, it is sunny.'' can use the oidful empty atom {\tt situation1\#Sunny()}.
Adding a descriptor (e.g., an independent slot), corresponding non-empty atoms
{\tt Sunny(place->NYC)}
and
{\tt situation1\#Sunny(place->NYC)} can be used.
}

\pagebreak
The {\bf first symbolic representation} of the entire iconic Fig.~\ref{TAexample} is shown as a 
PSOA RuleML KB of atomic ground (variableless) facts below such that the ground-atom colors correspond to the (hyper)arc colors in Fig.~\ref{TAexample}, except that gray for {\tt Top} is new (right-aligned ``\texttt{\%}'' comments indicate hierarchical structure):
\begin{small}
\begin{Verbatim}[commandchars=\\\{\}]
                                                      % (KB1)
Teacher##Scholar                                        % Taxonomy
Student##Scholar
TA##Teacher
TA##Student
\textcolor{brown}{John#TA(workload+>high)}                                 % Data (i) Fact 1
\textcolor{red}{John#Teacher(+[Wed Thu])}                                %          Fact 2
\textcolor{red}{John#Teacher(dept+>Physics)}                             %          Fact 3
\textcolor{red}{John#Teacher(salary+>29400)}                             %          Fact 4
\textcolor{ForestGreen}{John#Student(+[Mon Tue Fri])}                            %          Fact 5
\textcolor{ForestGreen}{John#Student(dept+>Math)}                                %          Fact 6
\textcolor{ForestGreen}{John#}\textcolor{gray}{Top}\textcolor{ForestGreen}{(-[1995 8 17])}   % \textcolor{gray}{Top} abstracted from \textcolor{ForestGreen}{Student}             Fact 7
\textcolor{ForestGreen}{John#}\textcolor{gray}{Top}\textcolor{ForestGreen}{(gender->male)}   % \textcolor{gray}{Top} abstracted from \textcolor{ForestGreen}{Student}             Fact 8
\textcolor{red}{John#}\textcolor{gray}{Top}\textcolor{red}{(income->29400)}  % \textcolor{gray}{Top} abstracted from \textcolor{red}{Teacher}             Fact 9
\end{Verbatim}
\end{small}

In {\tt (KB1)}'s upper four facts, representing the TA-diamond taxonomy part of Fig.~\ref{TAexample}, the ``\texttt{\#\#}'' infix indicates the binary subpredicate relation.

In Data (i) Facts 1 to 9, the data part constituting the rest of Fig.~\ref{TAexample} is represented
according to Section~\ref{intro}'s method (i) such that there are only single-descriptor atoms
(i.e., according to a
simplified version of Section~\ref{unscopingtodescribution}'s 
descributed normal form {\tt (KB3')}).
Particularly, in Data (i) Facts 7 to 9, the unique root predicate {\tt Top} is employed, which keeps this symbolic form of the method (i) representation unique
(by avoiding to choose from the non-{\tt Top} predicates).
These {\tt Top}-typed atoms can also be regarded as untyped atoms, as often used in F-logic and RIF.

The dual ``\texttt{+}'' vs. ``\texttt{-}'' marks are uniformly used for, respectively, dependent vs. 
independent descriptors, leading to four kinds of descriptors 
(exemplified with the descriptors of
some of
{\tt (KB1)}'s {\tt Student}
and {\tt Top}
atoms):
\begin{itemize}
\item For tuples, ``\texttt{+}'' vs. ``\texttt{-}'' are used as prefixes for the
square brackets, yielding the syntaxes \texttt{+[...]} vs. \texttt{-[...]}, e.g. \texttt{+[Mon Tue Fri]} vs. \texttt{-[1995 8 17]}.\footnote{In earlier PSOA versions, no prefix was used on any (square-)bracketed tuple,
and for atoms with an explicitly bracketed or a non-bracketed tuple dependency was decided on the basis of their predicate being ``relational'' \cite{ZouBoley16}.
Since PSOA 1.0, a prefix is used on every bracketed tuple, and a non-bracketed tuple is interpreted as dependent.}
\item For slots, ``\texttt{+}'' vs. ``\texttt{-}'' are used as shafts of the infix arrows, yielding the syntaxes \texttt{...+>...} vs. \texttt{...->...}, e.g. \texttt{dept+>Math} vs. \texttt{gender->male}.\footnote{In earlier PSOA versions, the ``\texttt{-}'' shaft was used for each arrow and the arrow-infixed slot was always interpreted as independent.}
\end{itemize}

Specifically, in each atom, a predicate such as {\tt Student}
-- besides typing a possible OID such as {\tt John} -- may be ``\texttt{(}$\dots$\texttt{)}''-applied to
one of the above dependent or independent descriptors (tuples or slots), e.g. {\tt Student} to {\tt dept+>Math}.

According to the metamodel of Appendix~\ref{sec:psoametamodel}, {\tt (KB1)}'s Data (i) Facts 1 to 6 \linebreak
-- all with a dependent descriptor -- are dependent atoms, while its Data (i) Facts 7 to 9 -- all with an independent descriptor -- are independent atoms.

\pagebreak
The {\bf second symbolic representation} of Fig.~\ref{TAexample} changes
(only) the data part
according to Section~\ref{intro}'s method (ii) such that there are a single- and two multiple-descriptor atoms complying to Fig.~\ref{TAexample}'s colors:
\vspace{-0.6em}
\mycomment{
\begin{small}
\begin{verbatim}
                                                      % (KB2)
. . .                                                   % Taxonomy
John#TA(workload+>high)                                 % Data (ii)
John#Teacher(+[Wed Thu] dept+>Physics salary+>29400 income->29400)
John#Student(+[Mon Tue Fri] -[1995 8 17] dept+>Math gender->male)
\end{verbatim}
\end{small}
}
\begin{small}
\begin{Verbatim}[commandchars=\\\{\}]
                                                      % (KB2)
. . .                                                   % Taxonomy
\textcolor{brown}{John#TA(workload+>high)}                                 % Data (ii)
\textcolor{red}{John#Teacher(+[Wed Thu] dept+>Physics salary+>29400 income->29400)}
\textcolor{ForestGreen}{John#Student(+[Mon Tue Fri] -[1995 8 17] dept+>Math gender->male)}
\end{Verbatim}
\end{small}
\vspace{-0.1em}

These lower three ground facts represent the data as dependence-concentrated atoms in the
-- logically immaterial --
color order ``brown-red-green''.
Such atoms can arbitrarily distribute independent descriptors, e.g. moving one to the {\tt TA} fact.

Generally, in each atom, a predicate
such as {\tt Student} 
-- besides typing (i.e., acting as a class of) a possible OID such as {\tt John} -- can be applied to zero or more dependent and independent descriptors (tuples and slots). Here, {\tt Student} is applied to four descriptors of all four kinds (in/dependent tuples/slots).

According to the metamodel of Appendix~\ref{sec:psoametamodel}, both of {\tt (KB2)}'s last two facts are independent+dependent psoa atoms.
Another case is independent psoa \linebreak
atoms, only having independent descriptors, e.g.
{\tt John\#Student(-[1995 8 17] gender->male)}.
These, then, further specialize to
psoa framepoints, independent atoms only having independent slots, e.g.
{\tt John\#Student(gender->male)}.
Such a framepoint atom corresponds to an F-logic-like typed frame, which is often -- e.g. in W3C RIF~\cite{BoleyKifer13} -- rewritten to
a conjunction of a membership and an untyped frame, e.g., in PSOA RuleML's presentation syntax,
{\tt And(John\#Student John\#Top(gender->male))}.
Similarly, such a typed frame that happens to have just one independent slot in RDF corresponds to a KB of a typing triple and one slot triple; e.g., the above framepoint in simplified N-Triples syntax becomes
\vspace{-0.4em}
\begin{small}
\begin{verbatim}
John rdf:type Student.
John gender male.
\end{verbatim}
\end{small}
\vspace{-0.2em}
An issue with triples and untyped frames is that, by detaching the predicate\footnote{PSOA's notion of `predicate' can be regarded as a generalization of, e.g., RDF's notion of `class'. However, RDF's notion of (binary/dyadic) `predicate' corresponds to
RIF's and PSOA's notion of `slot name'.} acting as a class from the OID, they cannot easily accommodate (predicate-) 
dependent slots, as provided, for example, by the special case of dependent psoa atoms that only have dependent slots, e.g.
{\tt John\#Student(dept+>Math)}.

{\tt (KB2)}'s data part distributes the independent descriptors
across
the {\tt Teacher} and {\tt Student} atoms, in one of several possible ways according to Section~\ref{intro}'s non-unique method (ii), where, e.g., the {\tt TA} atom could also receive one, two, or all three independent descriptors. In the unique \textbf{method (iii)} all independent descriptors are extracted from form (ii) and collected in one independent atom 
(using the unique root predicate {\tt Top}),\footnote{If a non-{\tt Top} predicate such as {\tt Teacher} were used for collecting all independents, the meaning would not change 
(all descriptors are independent of any predicate) but uniqueness would be lost. Additionally, for the uniqueness of such symbolic forms, a canonical order of
the bags of descriptors (tuples before slots, dependent before independent) and lexicographic order of the slots are normally used.} obtaining the following
unique {\it dependency-concentrated form} of the data, where the colors are like in {\tt (KB1)}:
\vspace{-0.1em}
\begin{small}
\begin{Verbatim}[commandchars=\\\{\}]
                                                      % (KB3)
. . .                                                   % Taxonomy
\textcolor{brown}{John#TA(workload+>high)}                                 % Data (iii)
\textcolor{red}{John#Teacher(+[Wed Thu] dept+>Physics salary+>29400)}
\textcolor{ForestGreen}{John#Student(+[Mon Tue Fri] dept+>Math)}
John#\textcolor{gray}{Top}(\textcolor{ForestGreen}{-[1995 8 17] gender->male} \textcolor{red}{income->29400})
\end{Verbatim}
\end{small}



Generally, a psoa atom with one or more dependent descriptors and no independent descriptor is called a {\it dependent atom}. If $\Pi$ is the predicate of a dependent atom (on which its descriptors are dependent), it is also called a {\it $\Pi$-dependent atom} (in Section \ref{rulesandqueries} this notion will be lifted to facts \& rules).
Complementarily, a psoa atom with one or more independent descriptors and no dependent descriptor is
an {\it independent atom}.
E.g., the first three atoms in {\tt (KB3)} are {\tt TA}-, {\tt Teacher}-, and {\tt Student}-dependent; the last atom is independent.

Posing ground queries to the ground atoms of {\tt (KB3)} exemplifies a \linebreak
{\it prerequisite for psoa-term unification}, which generalizes oidless-positional-term unification \cite{Lloyd87}
(this prerequisite applies also to non-ground atoms in queries and
KB clauses):
To unify, two psoa terms must ``pair up''~\cite{POSL10} descriptors
of the same dependency kind 
-- either both independent or both dependent -- \linebreak
after {\tt Top}-dependent 
descriptors have been reduced (Footnote~\ref{footnote:deptoindep}: ``toggled'') to \linebreak
independent descriptors. The following examples systematically change the \linebreak
dependency kind for slots and tuples in the KB and the query {\bf without using any {\tt Top}-dependent descriptors} 
(queries will be indicated by a ``{\tt >}~'' prompt):
\begin{scriptsize}
\begin{verbatim}
                                        % Slots
John#Student(... gender->male)            % Fragment of (KB2)
> John#Student(gender->male)
success
> John#Student(gender+>male)
fail     % "->" with "+>" violates same-dependency-kind prerequisite

John#Student(... dept+>Math ...)          % Fragment of (KB2)
> John#Student(dept->Math)
fail     % "+>" with "->" violates same-dependency-kind prerequisite
> John#Student(dept+>Math)
success
                                        % Tuples
John#Student(... -[1995 8 17] ...)        % Fragment of (KB2)
> John#Student(-[1995 8 17])
success
> John#Student(+[1995 8 17])
fail     % "-[" with "+[" violates same-dependency-kind prerequisite

John#Student(+[Mon Tue Fri] ...)          % Fragment of (KB2)
> John#Student(-[Mon Tue Fri])
fail     % "+[" with "-[" violates same-dependency-kind prerequisite
> John#Student(+[Mon Tue Fri])
success
\end{verbatim}
\end{scriptsize}
Here are examples {\bf with {\tt Top}-dependent descriptors in the KB, the query, or both}, hence performing ``(KB) toggling'' (lifting the ``toggled to'' notion from descriptors to their atoms and from atoms to their KBs), traced by indentation:
\begin{small}
\begin{verbatim}
                                        % Slots
John#Top(gender+>male)                    % (KB*)
 John#Top(gender->male)                   % Toggled (KB*)
> John#Student(gender->male)
fail     % Specific type for John not asserted but queried to be Student

John#Student(... gender->male)            % Fragment of (KB2)
> John#Top(gender+>male)
   John#Top(gender->male)
success  % Specific type for John asserted to be Student and not queried

John#Top(gender+>male)                    % (KB*)
 John#Top(gender->male)                   % Toggled (KB*)
> John#Top(gender+>male)
   John#Top(gender->male)
success  % Specific type for John not asserted and not queried
\end{verbatim}
\end{small}

For determining the above {\tt success} outcomes -- besides the same-dependency-kind prerequisite -- psoa unification, hence resolution, could be implemented (e.g., by generalizing OO~jDREW's POSL interpreters~\cite{POSL10} to PSOA), complementing \linebreak
PSOATransRun's PSOA translators as actually realized (Section~\ref{sec:implement}).
For example,
given
the KB 
{\tt John\#Student(+[Mon Tue Fri] ...)}, a dependency-agreeing non-ground 
(variable-containing) query {\tt John\#Student(+[Mon ?y ?z])} could apply unification to succeed with {\tt ?y}={\tt Tue} and {\tt ?z}={\tt Fri}.
But the PSOATransRun-realized prerequisite for psoa-term unification
allows
fast-failure decisions, \linebreak
as indicated by ``{\tt \%}'' comments in some of the above {\tt fail} outcomes.
Thus,
the \linebreak
dependency dimension
can support both 
expressivity and
efficiency.

Perspectival data, as in {\tt (KB2)}, are the basis of {\it perspectival querying}, as exemplified here with fixed ({\tt Teacher}, {\tt Student}) and variable ({\tt ?Persp}) perspectives:
\begin{small}
\begin{verbatim}
> John#Teacher(dept+>?unit)  % Under the perspective of John as a Teacher
?unit = Physics              % his department is Physics

> John#Student(dept+>?unit)  % Under the perspective of John as a Student
?unit = Math                 % his department is Math

> John#?Persp(dept+>?unit)   % Under the perspective of John as a ...
?Persp=Teacher ?unit=Physics % ... Teacher his department is Physics
?Persp=Student ?unit=Math    % ... Student his department is Math
\end{verbatim}
\end{small}
The predicate variable {\tt ?Persp} is bound non-deterministically by PSOATransRun.

\vspace{-0.7em}
\subsection{Possible Dependence-to-Independence Reductions}\label{dependencetranslations}

We now discuss possible reductions that translate dependent descriptors to independent descriptors, mainly by encoding the former as the latter.

Reductions of kinds of psoa atoms to other kinds
have already been done before the introduction of dimension D$_3$ for descriptor dependency
(cf. Appendix~\ref{sec:psoametamodel})
such as, in dimension D$_2$ for descriptor variety,
of a tuple to slots
(``positional-to-slotted'', with slot names like {\tt arg1}, ..., {\tt argN}) and vice versa (``slotted-to-positional'') of slots to a tuple \cite{BoleyShafiq+11}.
Both of these should now be done in a {\it dependency-preserving} manner, so that
an independent (resp., dependent) tuple reduces to a bag of independent (resp., dependent) slots,
and
a bag of independent (resp., dependent) slots reduces to an independent (resp., dependent) tuple.
Other reductions are likewise possible such as, in the OID dimension D$_1$, of oidful to oidless atoms (moving the OID to a new left-most argument position \cite{POSL10},
similarly as
on the runtime level by PSOATransRun's TPTP/Prolog primitives, cf. Section~\ref{sec:implement}) and vice versa (PSOA's objectification, cf. \cite{Boley11d,Boley15b,ZouBoley16}).

The current subsection augments these to considerations of reductions, in D$_3$, of
dependent to independent descriptors,
which could be complemented by reductions of
independent to dependent descriptors
(again, as done on the runtime level by PSOATransRun).
However, of all these reductions, only (static or dynamic \cite{ZouBoley16}) objectification of oidless to oidful atoms, in D$_1$, is
required by PSOATransRun (as will be indicated),
while
the reductions
in every dimension
contribute to maximum interoperability
with PSOA as a canonical language.

For dependent descriptors that are dependent slots a simple encoding is as follows.
For a pair of a predicate $p$ and a slot name $s$, a new slot name $s${\tt @}$p$ is introduced, where
``{\tt @}'' is assumed to be a reserved infix character indicating that the slot name is used `at' the predicate.
For example, the dependent atom
{\tt John\#Student(dept+>Math)}
would become the independent atom
{\tt John\#Student(dept@Student->Math)},
while
{\tt John\#Teacher(dept+>Physics)} \linebreak
would become
{\tt John\#Teacher(dept@Teacher->Physics)},
etc.
A disadvantage of this encoding is that, as one new name, $s${\tt @}$p$ is indivisible,
hence $s_1${\tt @}$p_1$ (e.g., {\tt dept@Student}) and $s_1${\tt @}$p_2$ (e.g., {\tt dept@Teacher}) appear as different as, say, $s_1${\tt @}$p_1$ and $s_2$ (e.g., {\tt income}).
A further
problem
is lack of scalability:
The combinatorics of concatenating\footnote{Since we use the abridged PSOA syntax, e.g. omitting indicators for local constants, we just need to concatenate the slot names. In the internal unabridged PSOA syntax, a slightly more involved combination of slot names would be needed.}
a slot name with (``{\tt @}'' and) predicates to form new slot names leads to
multiplicative growth
in the number of slot names, which creates issues for
KB interchange. In particular, for real-world applications, the slot name vocabulary (e.g., a subPropertyOf taxonomy) may well become unmanageable.

Another encoding would make use of slots as (syntactically) `higher-order' \linebreak
functions. 
For a pair of a predicate $p$ and a slot name $s$, a new complex slot name $s$($p$) is introduced, where the slot name s becomes a function taking the predicate $p$ (hence `higher-order') as the only argument.
For example, the dependent atom
{\tt John\#Student(dept+>Math)}
would become the independent atom
{\tt John\#Student(dept(Student)->Math)}.
This encoding would not have the vocabulary scalability problem since no new symbols are needed.
A problem is the encoding-caused transition from function-free (Datalog-like) PSOA RuleML languages
to function-using (Hornlog-like) ones, which are even `higher-order'.

A third conceivable, quite different, translation, basically employing a context for each perspective, will be discussed in Section~\ref{sec:conc}.

An obvious disadvantage of all these translations is the issue of unique inverse translation for reserved symbols such as ``{\tt @}'' and for encoding constructs such as complex terms like {\tt dept(Student)}.

For dependent descriptors that are dependent tuples, the situation is yet different.
One possibility would be reducing dependent tuples to dependent slots, as indicated in
Footnote~\ref{footnote:prop},
and then applying one of the above encodings (with their mentioned drawbacks).

Overall, since there
is no uniformly `best' translation and since 
dependence is the usual case for tuples, such as in relationships,
and for efficiency (cf. Section~\ref{sec:implement}),
we prefer to allow the direct modeling of dependent descriptors in the PSOA RuleML subfamily of languages (which still contains PSOA languages that do not make use of dependence but -- for modeling predicate-dependent knowledge -- \linebreak
would require some of the discussed dependence-reducing translations).

\subsection{Formal Rules and Their Querying}\label{rulesandqueries}

Let us now proceed to rules (in particular, implications): they can use non-ground versions of all four of the psoa descriptors
anywhere in their conclusion (head) and condition (body) atoms.
We will highlight the unusual cases of dependent slots,  \texttt{...+>...}, in {\tt (R1)}, and independent tuples, \texttt{-[...]}, in  {\tt (R2)}.

{\tt (KB1)}'s-{\tt (KB3)}'s {\tt John}-focused
dependent fact
{\tt John\#TA(workload+>high)} \linebreak
can be replaced by the following more versatile conclusion-dependent rule over dependent slots (and built-ins), where the fact's overall color, brown, is refined with a new color, orange, for the {\tt John}-generalizing variable {\tt ?o}:
\begin{small}
\begin{Verbatim}[commandchars=\\\{\}]
Forall ?o ?ht ?hs (            % (R1)
  \textcolor{orange}{?o}\textcolor{brown}{#TA(workload+>high)} :-
    And(\textcolor{orange}{?o}#Teacher(coursehours+>?ht)
        External(pred:numeric-greater-than(?ht 10))    % ?ht > 10
        \textcolor{orange}{?o}#Student(coursehours+>?hs)
        External(pred:numeric-greater-than(?hs 18)))   % ?hs > 18
)
\end{Verbatim}
\end{small}
The rule conclusion deduces -- for arbitrary OIDs {\tt ?o} that are member of {\tt TA} -- a {\tt TA}-dependent slot {\tt workload+>high} from a 
condition performing arithmetic threshold comparisons
for a {\tt Teacher}-dependent slot 
{\tt coursehours+>?ht}
and a {\tt Student}-dependent slot {\tt coursehours+>?hs}.
The three {\tt ?o} occurrences refer to the same individual, but under different perspectives.
The rule thus augments each condition-satisfying OID {\tt ?o} with the dependent qualitative {\tt workload} slot.

Assuming that {\tt (KB1)}'s-{\tt (KB3)}'s {\tt Teacher}/{\tt Student} descriptors for {\tt John} are augmented by corresponding dependent quantitative {\tt coursehours} slots,
\begin{small}
\begin{verbatim}
John#Teacher(... coursehours+>12 ...)
John#Student(... coursehours+>20 ...)
\end{verbatim}
\end{small}
the combined changes for, e.g., {\tt (KB2)} lead to what is called {\tt (KB2\#)} in Section~\ref{sec:presentationsyntax}, and adding the rule {\tt (R1)} we arrive at a sample KB that is called {\tt (KB)} in Fig.~\ref{SampleKB}.
The rule successfully answers the following dependent-slot (``{\tt +}'') ground query:
\begin{small}
\begin{verbatim}
> John#TA(workload+>high)                      % (Q+1)
\end{verbatim}
\end{small}
For this, the query is first unified with the conclusion, with the internal binding {\tt ?o} = {\tt John}.
Then, in the condition, the first/third conjunct performs a ``look-in''-retrieval~\cite{Boley15b} of the {\tt \_Teacher}/{\tt \_Student}-dependent 
{\tt \_coursehours} {\tt 12}/{\tt 20} slot ``in'' (i.e., as part of) the corresponding fact;
the second/fourth conjunct ``{\tt >}''-compares the {\tt \_coursehours} filler with its threshold {\tt 10}/{\tt 18}.
The (RIF-like) {\tt External} wrapper
is employed here, as usually,  for built-in calls.

Similarly, the rule makes the dependent-slot non-ground (``{\tt ?}'') query
\begin{small}
\begin{verbatim}
> ?who#TA(workload+>?level)                    % (Q+1?)
\end{verbatim}
\end{small}
succeed, with bindings {\tt ?who} = {\tt John} and {\tt ?level} = {\tt high}.\footnote{Besides the {\tt TA}-dependent {\tt workload} being defined here via a double threshold of {\tt Teacher}- and {\tt Student}-dependent {\tt coursehours}, rules for {\tt Teacher}- and {\tt Student}-dependent {\tt workload}s could also be defined, e.g.: {\tt ?o\#Teacher(workload+>high) :- And(?o\#Teacher(coursehours+>?ht) External(pred:numeric-greater-than(?ht 16)))}. Since {\tt John}'s {\tt 12} {\tt Teacher}-dependent {\tt coursehours} are not greater than this rule's threshold of {\tt 16}, a {\tt Teacher}-dependent query {\tt John\#Teacher(workload+>high)} would fail, unlike the {\tt TA}-dependent queries.}

A conclusion-independent rule mapping from a ({\tt ValidDate}d) independent tuple to independent slots can be used to test whether the three elements of the tuple
constitute a valid date
and putting such elements into the filler positions of
appropriately
named slots:
\mycomment{
\begin{small}
\begin{verbatim}
Forall ?o ?y ?m ?d (                    % (R2)
  ?o#Person(year->?y month->?m day->?d) :-
    And(?o#Person(-[?y ?m ?d]) Year(?y) Month(?m) DayOfMonthOfYear(?d ?m ?y))
)
\end{verbatim}
\end{small}
}
\begin{small}
\begin{verbatim}
Forall ?o ?y ?m ?d (                       % (R2)
  ?o#Person(year->?y month->?m day->?d) :-
    And(?o#Person(-[?y ?m ?d]) ValidDate(?y ?m ?d))
)
\end{verbatim}
\end{small}
The rule thus enriches an OID {\tt ?o} of predicate {\tt Person} that is described with such a tuple by the three slots {\tt year}, {\tt month}, and {\tt day}.

We assume that {\tt (KB1)}-{\tt (KB3)} are augmented by {\tt (R2)} as well as the following subpredicate fact and {\tt ValidDate}-checking rule\footnote{The ``{\tt ...}'' conjuncts stand for subrule queries ensuring, e.g., 28 days for February, except 29 in leap years. Finite subsets of triples from the infinite virtual date table, including {\tt ValidDate(1995 8 17)}, could also be 
materialized as facts.}:

\mycomment{
ground and non-ground (universally quantified) facts like
\begin{small}
\begin{verbatim}
Scholar##Person    % Extend TA diamond by a new subtaxonomy root
Year(1995)                   % ?y < 0 OR 0 < ?y
Month(8)                     % 1 =< ?m =< 12
Forall ?y (DayOfMonthOfYear(17 8 ?y))    % ?d = 28, 29, 30, or 31 for given ?m and ?y
\end{verbatim}
\end{small}
or rules like
{\small\tt Month(?m)}$\,${\small\tt :-}$\;${\small\tt And(External(pred:numeric-less-than-or-equal(1 ?m)) External(pred:numeric-less-than-or-equal(?m 12)))}
generalizing such facts \linebreak
(as indicated by ``{\tt \%}'' comments, where {\tt ?y} must be an integer while {\tt ?m} and {\tt ?d} must be positive integers),
}

\begin{small}
\begin{verbatim}
Scholar##Person              % Extend TA diamond by a new subtaxonomy root
Forall ?y ?m ?d ( ValidDate(?y ?m ?d) :- And(...) )  % Ensure date triples
\end{verbatim}
\end{small}

Now, rule {\tt (R2)} successfully answers the independent-slot (``{\tt -}'') ground query
\begin{small}
\begin{verbatim}
> John#Person(year->1995 month->8 day->17)                    % (Q-2)
\end{verbatim}
\end{small}
and succeeds for the independent-slot non-ground query
\begin{small}
\begin{verbatim}
> John#Person(year->?ye month->?mo day->?da)                  % (Q-2?)
\end{verbatim}
\end{small}
with bindings {\tt ?ye} = {\tt 1995}, {\tt ?mo} = {\tt 8}, and {\tt ?da} = {\tt 17}.

\section{Equivalence Laws for PSOA Knowledge}\label{lawsknowtrans}

In this section we continue the discussion of Section~\ref{foundknowrep} about knowledge representation in PSOA by explaining laws used for its knowledge transformation, namely unscoping, describution and centralization, rescoping, as well as default expansion.
The laws are formalized as meta-level equivalences (``$\equiv$'') usable left (top) to right (bottom) and  right to left.
As equivalence laws, they define (semantics-preserving) equivalence classes of formulas, thus further preparing the model-theoretic semantics in Section~\ref{sec:semantics}.
Some of these equivalences will also be taken up -- used in one direction, for non-empty atoms -- for the translation-based implementation in Section~\ref{sec:implement}.
In the following subsections we assume oidful atoms (oidless atoms require prior objectification),
except for Section~\ref{sec:inheritance}, where oidless facts are expanded into oidful rules (and vice versa).

\subsection{From Unscoping to Describution and Centralization}\label{unscopingtodescribution}
In this subsection we discuss unscoping and describution as well as centralization as the inverse~\cite{Boley15b} of describution.
For this, recall that independent descriptors are not sensitive to any specific (non-{\tt Top}) predicate in whose scope they occur within an atom. 

{\it Unscoping} of the independent descriptors in an independent atom with a non-{\tt Top} predicate extracts the atom's membership,
leaving behind an atom in which the non-{\tt Top} predicate is replaced by {\tt Top}.

Unscoping has
the following
general form,
where $\Omega$, $\Pi$ ($\neq$ {\tt Top}), and $\Delta^{\Mathtt{-}}_i$ are meta\-variables for, respectively,
OIDs, predicates, and independent ($^{-}$) 
descriptors
($s$ $\geq$ 0, where $s$ = 0, for empty atoms, is included for generality):
\[
\begin{array}{r@{}l}
\Omega\Mathtt{\#}\Pi\Mathtt{(} & 
\Delta^{\Mathtt{-}}_1 \sldots \Delta^{\Mathtt{-}}_s \Mathtt{)} \\
& \equiv \\
\Mathtt{And (}\Omega\Mathtt{\#}\Pi{}\  \Omega\Mathtt{\#} & \Mathtt{Top}\Mathtt{(}\Delta^{\Mathtt{-}}_1\sldots\Delta^{\Mathtt{-}}_s \Mathtt{)} \Mathtt{)}
\end{array}
\]

Describution, which has also been called tupribution/slotribution and will be further characterized in the second half of this subsection,
is similar to
unscoping
but decomposes
a given zero-or-more-descriptor atom into a conjunction of single-descriptor atoms, where the given atom's OID is `distributed' over the conjoined atoms with their single descriptors (tuples or slots).

Next, we develop examples for unscoping and describution as applied to queries, facts, and rules.

For instance, complementing the ground-query dependent-slot atom {\tt (Q+1)} and the non-ground-query dependent-slot atom {\tt (Q+1?)} of Section \ref{rulesandqueries}, their dual ground and non-ground independent-slot queries are (with ``\texttt{+>}'' toggled to ``\texttt{->}''):
\begin{small}
\begin{verbatim}
> John#TA(workload->high)                               % (Q-1)
> ?who#TA(workload->?level)                             % (Q-1?)
\end{verbatim}
\end{small}
Being (predicate-)independent, the slots of these two atoms can be unscoped \linebreak
-- from the predicate {\tt TA} to the vacuous predicate {\tt Top}, yielding untyped atoms -- 
by extracting the memberships {\tt John\#TA} and {\tt ?who\#TA} into separate conjuncts. By leaving behind {\tt John\#Top} and {\tt ?who\#Top}, {\tt Top} occurrences are introduced for unscoping, thus transforming the above atoms (here, the queries 
{\tt (Q-1)} and {\tt (Q-1?)})
to these equivalent conjunctions (here, conjunctive queries):
\begin{small}
\begin{verbatim}
And(John#TA John#Top(workload->high))                    % (C-1)
And(?who#TA ?who#Top(workload->?level))                  % (C-1?)
\end{verbatim}
\end{small}
Since {\tt (Q-1)} and {\tt (Q-1?)} already have single descriptors, {\tt (C-1)} and {\tt (C-1?)} are also their describution results.

While ({\tt ?who} =) {\tt John} is a {\tt TA} and as a {\tt TA} was deduced, in Section \ref{rulesandqueries}, by the rule {\tt (R1)} to have ({\tt ?level} =) {\tt high} {\tt workload},
generally, as a member of {\tt Top},
which is made explicit by unscoping, {\tt John} cannot be deduced by {\tt (R1)} to have any description,
because {\tt (R1)}'s conclusion retains the corresponding OID variable {\tt ?o} as a member of {\tt TA}.
This difference is due to the descriptor being independent in the query (leading to {\tt Top}) while being dependent in the rule conclusion (retaining the non-{\tt Top} predicate), so that the
prerequisite for psoa-term unification
of Section~\ref{factsandqueries}
is not fulfilled.
Therefore, the {\tt (C-1)} and {\tt (C-1?)} conjunctions (here, queries) fail.

As another example, refining the {\tt Person} predicate of the ground independent-slot query {\tt (Q-2)} and the non-ground independent-slot query {\tt (Q-2?)} of Section \ref{rulesandqueries}, their {\tt TA}-predicate versions are:
\begin{small}
\begin{verbatim}
> John#TA(year->1995 month->8   day->17)                  % (Q-3)
> John#TA(year->?ye  month->?mo day->?da)                 % (Q-3?)
\end{verbatim}
\end{small}

On one hand, unscoping of the atoms of queries {\tt (Q-3)} and {\tt (Q-3?)}
creates conjunctions where the membership {\tt John\#TA} is extracted and the atoms'  predicate {\tt TA} is evacuated to {\tt Top}:
\begin{small}
\begin{verbatim}
And(John#TA John#Top(year->1995 month->8   day->17))
And(John#TA John#Top(year->?ye  month->?mo day->?da))
\end{verbatim}
\end{small}

On the other hand, describution (here pure slotribution rather than pure tupribution or combined tupribution+slotribution) of the same atoms
creates conjunctions where the membership {\tt John\#TA} is extracted and all (here, three) slots are used for {\tt Top}-typed single-slot atoms:
\begin{small}
\begin{verbatim}
And(John#TA John#Top(year->1995) John#Top(month->8)   John#Top(day->17))
And(John#TA John#Top(year->?ye)  John#Top(month->?mo) John#Top(day->?da))
\end{verbatim}
\end{small}

Again, {\tt John\#TA} can be shown by fact retrieval; the conclusion of the rule {\tt (R2)} of Section \ref{rulesandqueries} is also transformed by slotribution, so that the entire conjunctions, hence {\tt (Q-3)} and {\tt (Q-3?)}, can be successfully deduced with the same answers as for {\tt (Q-2)} and {\tt (Q-2?)}.

The meta-level equivalence for {\it describution} (when used left to right) and {\it centralization} (when used right to left) has the following general form, 
where
$\Delta^{\Mathtt{+}}_i$ and $\Delta^{\Mathtt{-}}_j$ are names for, respectively, arbitrary dependent (``$^{\Mathtt{+}}$'') and independent
(``$^{-}$'') descriptors
($r$ $\geq$ 0, $s$ $\geq$ 0):\footnote{Note that the ``$^{\Mathtt{+}}$''/``$^{-}$'' superscripts -- like the subscripts -- are part of the metavariable names.
A unary prefix operator ``$\pm$'' can be used to toggle a dependent to an independent descriptor and vice versa, keeping its content unchanged.
It is defined with four equations on the concrete-descriptor level:
$\pm(${\tt +[}t\tsub{1}$\sldots$t\tsub{n}{\tt ]}$)$ $=$ {\tt -[}t\tsub{1}$\sldots$t\tsub{n}{\tt ]},
$\pm(${\tt -[}t\tsub{1}$\sldots$t\tsub{n}{\tt ]}$)$ $=$ {\tt +[}t\tsub{1}$\sldots$t\tsub{n}{\tt ]},
$\pm($p{\tt +>}v$)$ $=$ p{\tt ->}v,
$\pm($p{\tt ->}v$)$ $=$ p{\tt +>}v.
For any descriptor $\Delta$, $\pm(\pm(\Delta))$ $=$ $\Delta$.
The prefix ``$\pm$'' can be applied (omitting the parentheses)
on the right-hand side of a meta-level equivalence
between atoms with $r$ descriptors that are marked as dependent on {\tt Top}, and their togglings, which are marked as independent:
$\Omega\Mathtt{\#}\Mathtt{Top}\Mathtt{(}\Delta^{\Mathtt{+}}_1\sldots \Delta^{\Mathtt{+}}_r \ \Delta^{\Mathtt{-}}_1 \sldots \Delta^{\Mathtt{-}}_s \Mathtt{)}
\equiv
\Omega\Mathtt{\#}\Mathtt{Top}\Mathtt{(}\!\pm\!\Delta^{\Mathtt{+}}_1\sldots \pm\!\Delta^{\Mathtt{+}}_r \ \Delta^{\Mathtt{-}}_1 \sldots \Delta^{\Mathtt{-}}_s \Mathtt{)}$.
\label{footnote:deptoindep}}
\[
\begin{array}{rl}
\Omega\Mathtt{\#}\Pi\Mathtt{(} 
\Delta^{\Mathtt{+}}_1\sldots \Delta^{\Mathtt{+}}_r & \Delta^{\Mathtt{-}}_1 \sldots \Delta^{\Mathtt{-}}_s \Mathtt{)} \\
\equiv \\
\Mathtt{And (}\Omega\Mathtt{\#}\Pi{}\ \Omega\Mathtt{\#}\Pi\Mathtt{(}\Delta^{\Mathtt{+}}_1 \Mathtt{)}
\sldots
\Omega\Mathtt{\#}\Pi\Mathtt{(}\Delta^{\Mathtt{+}}_r \Mathtt{)}
&
\Omega\Mathtt{\#}\Mathtt{Top}\Mathtt{(}\Delta^{\Mathtt{-}}_1 \Mathtt{)}
\sldots
\Omega\Mathtt{\#}\Mathtt{Top}\Mathtt{(}\Delta^{\Mathtt{-}}_s \Mathtt{)}
\Mathtt{)}
\end{array}
\]

\vspace{0.6em}  
The general case of describution will be further explained on the concrete-descriptor level in Section~\ref{sec:implement}. It corresponds to Section~\ref{intro}'s method (i), transforming a zero-or-more-descriptor atom into a conjunction of one membership and zero or more single-descriptor atoms, 
where each independent descriptor's
non-{\tt Top} predicate is evacuated to {\tt Top},
as in unscoping, while each dependent descriptor is kept within the scope of the original predicate.

Describution is applicable to each atom of a query or a KB.
For example, the
three
facts of {\tt (KB2)} can be transformed to this {\it descributed normal form} \linebreak
(pretty-printed so that the same kinds of descriptors are on the same line):
\begin{small}
\begin{verbatim}

                                                      % (KB2')
. . .                                                   % Taxonomy
And(John#TA John#TA(workload+>high))                    % Data (ii')
And(
    John#Teacher 
    John#Teacher(+[Wed Thu])
    John#Teacher(dept+>Physics) John#Teacher(salary+>29400)
    John#Top(income->29400)
)
And(
    John#Student
    John#Student(+[Mon Tue Fri])
    John#Top(-[1995 8 17])
    John#Student(dept+>Math)
    John#Top(gender->male)
)
\end{verbatim}
\end{small}

The conjuncts can be regrouped to collect all independent descriptors into a separate conjunction
(pretty-printed as above), which is also the descributed normal form of {\tt (KB3)}:

\begin{small}
\begin{verbatim}
                                                     % (KB3')
. . .                                                   % Taxonomy
And(John#TA John#TA(workload+>high))                    % Data (iii')
And(
    John#Teacher 
    John#Teacher(+[Wed Thu])
    John#Teacher(dept+>Physics) John#Teacher(salary+>29400)
)
And(
    John#Student
    John#Student(+[Mon Tue Fri])
    John#Student(dept+>Math)
)
And(
    John#Top(-[1995 8 17])
    John#Top(gender->male) John#Top(income->29400)
)
\end{verbatim}
\end{small}

This shows the logical equivalence between {\tt (KB2)} and its dependency-concen\-trated form {\tt (KB3)}.

The conjuncts can also be directly used in the (implicit) top-level conjunction of (the {\tt Assert} of) a PSOA RuleML KB (cf. Section~\ref{sec:presentationsyntax}).

\subsection{Rescoping as Describution and Centralization}

Building on Section~\ref{unscopingtodescribution}, we now explain rescoping for oidful atoms. {\it Rescoping} removes an independent descriptor of
an
atom that has some OID and predicate and adds this independent descriptor to
an
atom that has the same OID but in the non-trivial case has a different predicate.
In the taxonomy, the two predicates may
(a) be on the same taxonomic level -- i.e., have an equal shortest distance to the root predicate {\tt Top} -- 
(``horizontal'' rescoping), 
(b) be on the same taxonomic inheritance line 
-- i.e., be on the same path to {\tt Top} -- (``vertical'' rescoping),
or 
(c) be taxonomically unrelated -- i.e., neither (a) nor (b) applies -- (``diagonal'' rescoping).
Rescoping first does unscoping (for a single-descriptor atom) or, generally, describution (for a one-or-more-descriptor atom); it then does centralization,
targeting the (scope of the) other predicate.

Rescoping has the following general form ($r$ $\geq$ 0 and $r'$ $\geq$ 0 because there need not be any dependent descriptor, $s$ $\geq$ 0 because there need not be any independent descriptor in the rescoping target, and $s'$ $\geq$ 1 because there must be at least the independent descriptor $\Delta'^{\Mathtt{-}}_{i'}$ in the rescoping source):\footnote{An enriched form of rescoping could be introduced, where one or more independent descriptors are moved to the rescoping target together.
However, such multi-descriptor rescoping can be reduced to repeated single-descriptor rescopings.}

\mycomment{
\begin{align*}
\Mathtt{And (} \Omega\Mathtt{\#}\Pi\Mathtt{(} \Delta^{\Mathtt{+}}_1\sldots \Delta^{\Mathtt{+}}_r & \  \Delta^{\Mathtt{-}}_1\sldots \Delta^{\Mathtt{-}}_{i} \Delta^{\Mathtt{-}}_{i+1} \sldots \Delta^{\Mathtt{-}}_s \Mathtt{)} \\
\Omega\Mathtt{\#}\Pi'\Mathtt{(} \Delta'^{\Mathtt{+}}_1\sldots \Delta'^{\Mathtt{+}}_{r'} & \  \Delta'^{\Mathtt{-}}_1\sldots \Delta'^{\Mathtt{-}}_{i'-1} \Delta'^{\Mathtt{-}}_{i'} \Delta'^{\Mathtt{-}}_{i'+1}\sldots \Delta'^{\Mathtt{-}}_{s'} \Mathtt{)} \Mathtt{)} \\
& \equiv \\
\Mathtt{And (}\Omega\Mathtt{\#}\Pi\Mathtt{(} \Delta^{\Mathtt{+}}_1\sldots \Delta^{\Mathtt{+}}_r & \  \Delta^{\Mathtt{-}}_1\sldots \Delta^{\Mathtt{-}}_{i} \Delta'^{\Mathtt{-}}_{i'} \Delta^{\Mathtt{-}}_{i+1}\sldots \Delta^{\Mathtt{-}}_s \Mathtt{)} \\
\Omega\Mathtt{\#}\Pi'\Mathtt{(} \Delta'^{\Mathtt{+}}_1\sldots \Delta'^{\Mathtt{+}}_{r'} & \  \Delta'^{\Mathtt{-}}_1\sldots \Delta'^{\Mathtt{-}}_{i'-1} \Delta'^{\Mathtt{-}}_{i'+1} \sldots \Delta'^{\Mathtt{-}}_{s'} \Mathtt{)} \Mathtt{)}
\end{align*}
}

\[
\begin{array}{r@{}l@{}l@{\ }l}
\Mathtt{And (} & \Omega\Mathtt{\#}\Pi & \Mathtt{(} \Delta^{\Mathtt{+}}_1\sldots \Delta^{\Mathtt{+}}_r & 
\Delta^{\Mathtt{-}}_1\sldots \Delta^{\Mathtt{-}}_{i} \sldots \Delta^{\Mathtt{-}}_s \Mathtt{)} \\
& \Omega\Mathtt{\#}\Pi' & \Mathtt{(} \Delta'^{\Mathtt{+}}_1\sldots \Delta'^{\Mathtt{+}}_{r'} & \Delta'^{\Mathtt{-}}_1\sldots \Delta'^{\Mathtt{-}}_{i'-1} \Delta'^{\Mathtt{-}}_{i'} \sldots \Delta'^{\Mathtt{-}}_{s'} \Mathtt{)} \Mathtt{)} \\
& & & \equiv \\
\Mathtt{And (} & \Omega\Mathtt{\#}\Pi & \Mathtt{(} \Delta^{\Mathtt{+}}_1\sldots \Delta^{\Mathtt{+}}_r &
\Delta^{\Mathtt{-}}_1\sldots \Delta^{\Mathtt{-}}_{i} \Delta'^{\Mathtt{-}}_{i'} \sldots \Delta^{\Mathtt{-}}_s \Mathtt{)} \\
& \Omega\Mathtt{\#}\Pi' & \Mathtt{(} \Delta'^{\Mathtt{+}}_1\sldots \Delta'^{\Mathtt{+}}_{r'} 
& \Delta'^{\Mathtt{-}}_1\sldots \Delta'^{\Mathtt{-}}_{i'-1} \sldots \Delta'^{\Mathtt{-}}_{s'} \Mathtt{)} \Mathtt{)}
\end{array}
\]

For example,
assuming the ground facts of the example of Fig.~\ref{TAexample} are asserted as in Section \ref{factsandqueries},  {\tt (KB2)},
the below simple, ``horizontal'' rescoping in a conjunctive ground query uses
unscoping
of a 
{\tt Student}-independent slot from \linebreak
the {\tt Student} scope
followed by centralization targeting the {\tt Teacher} scope \linebreak
(intermediate derivation steps are traced using indentation):\footnote{To emphasize the target of the rescoped descriptor, we retain here the empty parentheses of {\tt John\#Teacher()} instead of using the shortened {\tt John\#Teacher}.}
\begin{small}
\begin{verbatim}
> And(John#Teacher() John#Student(income->29400))
\end{verbatim}
\end{small}
\vspace{-1.6em}
\begin{scriptsize}
\begin{verbatim}
    And(John#Teacher() And(John#Student() John#Top(income->29400)))
\end{verbatim}
\end{scriptsize}
\vspace{-1.8em}
\begin{footnotesize}  
\begin{verbatim}
   And(John#Teacher() John#Student() John#Top(income->29400))
   And(John#Teacher() John#Top(income->29400) John#Student())
\end{verbatim}
\end{footnotesize}
\vspace{-1.7em}
\begin{scriptsize}
\begin{verbatim}
    And(And(John#Teacher() John#Top(income->29400)) John#Student()))
\end{verbatim}
\end{scriptsize}
\vspace{-1.7em}
\begin{small}
\begin{verbatim}
   And(John#Teacher(income->29400) John#Student())
success
\end{verbatim}
\end{small}

Similarly,
using the same {\tt (KB2)},
the below crosswise,  ``horizontal'' rescopings of a conjunctive ground query use
`parallel' unscopings
of a {\tt Teacher}-independent tuple from the {\tt Teacher} scope
and a 
{\tt Student}-independent slot from the {\tt Student} scope
followed by two `parallel' centralizations targeting, respectively, the {\tt Student} and the {\tt Teacher} scope:
\pagebreak  
\begin{small}
\begin{verbatim}
> And(John#Teacher(-[1995 8 17]) John#Student(income->29400))
\end{verbatim}
\end{small}
\begin{scriptsize}
\begin{verbatim}
    And(And(John#Teacher John#Top(-[1995 8 17])) And(John#Student John#Top(income->29400)))
\end{verbatim}
\end{scriptsize}
\begin{footnotesize}  
\begin{verbatim}
   And(John#Teacher John#Top(-[1995 8 17]) John#Student John#Top(income->29400))
   And(John#Teacher John#Top(income->29400) John#Student John#Top(-[1995 8 17]))
\end{verbatim}
\end{footnotesize}
\begin{scriptsize}
\begin{verbatim}
    And(And(John#Teacher John#Top(income->29400)) And(John#Student John#Top(-[1995 8 17])))
\end{verbatim}
\end{scriptsize}
\begin{small}
\begin{verbatim}
   And(John#Teacher(income->29400) John#Student(-[1995 8 17]))
success
\end{verbatim}
\end{small}

Again
based on {\tt (KB2)}'s {\tt Teacher}-independent {\tt John\#Teacher} descriptors and {\tt Student}-independent {\tt John\#Student} descriptors,
the below multiway, ``vertical'' rescopings of a conjunctive ground query containing an atom with two {\tt TA}-independent {\tt John\#TA} descriptors use tupribution/slotribution-combining 
describution --
where the non-{\tt Top} predicate {\tt TA} is evacuated to {\tt Top},
as in unscoping --
followed by two `parallel' centralizations:
\begin{small}
\begin{verbatim}
> And(John#Teacher John#TA(-[1995 8 17] income->29400) John#Student)
\end{verbatim}
\end{small}
\begin{scriptsize}
\begin{verbatim}
    And(John#Teacher And(John#TA John#Top(-[1995 8 17]) John#Top(income->29400)) John#Student)
\end{verbatim}
\end{scriptsize}
\begin{scriptsize}
\begin{verbatim}
    And(John#Teacher John#TA John#Top(-[1995 8 17]) John#Top(income->29400) John#Student)
\end{verbatim}
\end{scriptsize}
\begin{scriptsize}
\begin{verbatim}
    And(John#TA John#Teacher John#Top(income->29400) John#Student John#Top(-[1995 8 17]))
\end{verbatim}
\end{scriptsize}
\begin{scriptsize}
\begin{verbatim}
    And(John#TA And(John#Teacher John#Top(income->29400)) And(John#Student John#Top(-[1995 8 17])))
\end{verbatim}
\end{scriptsize}
\begin{small}
\begin{verbatim}
   And(John#TA John#Teacher(income->29400) John#Student(-[1995 8 17]))
success
\end{verbatim}
\end{small}

Note that, although {\tt TA} is a subpredicate of both {\tt Teacher} and {\tt Student}, the derivation does not require this taxonomic information, but instead directly uses the multi-memberships of {\tt John} in the three predicates ({\tt John\#TA}, {\tt John\#Teacher}, and {\tt John\#Student}).

In contrast to an independent descriptor, the scope of a dependent descriptor is limited to the predicate of its enclosing atom, and no rescoping is allowed.
The below query-answer pairs exemplify, also based on {\tt (KB2)}:
\begin{small}
\begin{verbatim}
> And(John#Teacher(+[Mon Tue Fri]) John#Student)
fail    % Query tuple dependent on Teacher, rescoping for (KB2) impossible

> And(John#Teacher(+[Wed Thu]) John#Student)
success % Query tuple dependent on Teacher, rescoping for (KB2) unnecessary

> And(John#Teacher John#Student(dept+>Physics))
fail    % Query slot dependent on Student, rescoping for (KB2) impossible

> And(John#Teacher John#Student(dept+>Math))
success % Query slot dependent on Student, rescoping for (KB2) unnecessary
\end{verbatim}
\end{small}

Note that the unscoping of a single-independent-descriptor atom is equivalent to its describution,
which -- when the atom with OID $o$ is equivalently extended to a conjunction by a trivially true same-OID empty atom with predicate {\tt Top} of the form $o${\tt \#Top()} -- is a special case of rescoping the descriptor to $o${\tt \#Top()}.
For example, revisiting Section~\ref{unscopingtodescribution}, {\tt (Q-1)} and {\tt (Q-1?)} are equivalent to
\vspace{-0.2em}
\begin{small}
\begin{verbatim}
> And(John#TA(workload->high) John#Top())                   % (Q-1')
> And(?who#TA(workload->?level) ?who#Top())                 % (Q-1?')
\end{verbatim}
\end{small}
\vspace{-0.2em}
where the descriptors {\tt workload->high} and {\tt workload->?level} can be rescoped from the original atoms to the empty atoms, obtaining
Section~\ref{unscopingtodescribution}'s {\tt (C-1)} and {\tt (C-1?)}, again usable as queries.

\subsection{Default Descriptors and Their Inheritance}\label{sec:inheritance}

In AI knowledge representation, so-called ``default values'' (in PSOA: {\it default fillers}) permit {\it default slots} (names with fillers) to be inherited from a class to all of its instances.
PSOA RuleML allows a monotonic version of such inheritance also for {\it default tuples}, arriving at the generalized notion of {\it default descriptors}.

For realizing default-descriptor inheritance, monotonic {\it default rules} are used, which are rules whose conclusion derives descriptors for a universally quantified OID
from a condition that proves an OID-predicate membership,
where the OID represents all of the predicate's instances.
This proof may directly retrieve a membership (`base case') or proceed through one or more subpredicate facts to chain to a less general predicate (`recursive case').

Following the orthogonality principle, the
initial predicate %
is just the {\it seed} of the descriptors that are inherited to all of its instances --
the descriptors need \linebreak
not be dependent but can be
independent from their seed predicate. \linebreak
While dependent-descriptor default rules use a non-{\tt Top} conclusion predicate, namely the same predicate as in the condition, independent-descriptor default rules use the {\tt Top} conclusion predicate.

For example, {\tt (KB1)}-{\tt (KB3)} can be augmented by two independent-descriptor default rules as follows (cf. Int'l Standard Classification of Occupations\footnote{\url{http://www.ilo.org/public/english/bureau/stat/isco}}):
\vspace{-0.2em}
\begin{small}
\begin{verbatim}
Forall ?o (
  ?o#Top(-[2 3]  % ISCO major (2: Professionals) and sub-major (3: Teaching)
         offer->service
         aptitude->explanation) :-
    ?o#Teacher
)
\end{verbatim}
\end{small}
\vspace{-1.2em}
\begin{small}
\begin{verbatim}
Forall ?o (
  ?o#Top(acquire->KSAs  % Knowledge, Skills and Abilities
         aptitude->comprehension) :-
    ?o#Student
)
\end{verbatim}
\end{small}
\vspace{-0.3em}

Here are query examples inheriting default tuples and slots, where the default slot {\tt aptitude} becomes multi-valued with non-conflicting fillers {\tt explanation} and  {\tt comprehension} (for a conflicting example, see Section~\ref{sec:conc}):
\begin{small}
\begin{verbatim}
> John#Teacher(-[2 3] offer->service)
success
> John#Student(acquire->KSAs aptitude->?w)
?w=explanation    % Independently from Teacher
?w=comprehension  % Independently from Student
\end{verbatim}
\end{small}

These same answers are still obtained after removing the memberships \linebreak
{\tt John\#Teacher} and
{\tt John\#Student}
from {\tt (KB1)}-{\tt (KB3)},
since the remaining membership
{\tt John\#TA}
is reached from both rule conditions by one step of subpredicate chaining.

Moreover, if {\tt (KB1)}-{\tt (KB3)} are further augmented by the ground fact \linebreak
{\tt John\#TA(aptitude->illustration)},
a query like
{\tt John\#TA(aptitude->?w)}
exemplifies that PSOA uses -- to keep the semantics simple -- non-overriding, monotonic fillers, cumulatively binding {\tt ?w} to multiple values, {\tt illustration}, {\tt explanation}, and {\tt comprehension}.\footnote{For cases where it is preferable to regard {\tt illustration} as an `exception' overriding the other two default fillers, a non-monotonic semantics -- as, e.g., in Flora-2~\cite{flora2-manual} -- \linebreak
could be orthogonally added to PSOA, both for independent and dependent descriptors,
and selectively for certain slot names (e.g., {\tt policy} but not {\tt aptitude}), predicates, or slot-name-(dependent-on-)predicate combinations -- rather than for an entire KB.}

\mycomment{
Default rules can be regarded as {\it default facts} of the form 
$\Pi\Mathtt{\{}\Delta^{\Mathtt{+}}_1\sldots \Delta^{\Mathtt{+}}_r \  \Delta^{\Mathtt{-}}_1\sldots \Delta^{\Mathtt{-}}_s \Mathtt{\}}$, 
built from a new kind of atomic formula  -- indicated by parentheses replaced with curly braces -- retaining an {\tt ?o}-free version of the conclusion that acquires the condition predicate while omitting the condition itself.
For independent, slotted descriptors, PSOA's default facts are similar to Flora-2's~\cite{flora2-manual}
class-wide frame formulas 
$\Pi\Mathtt{[|}\Delta^{\Mathtt{-}}_1\sldots \Delta^{\Mathtt{-}}_s \Mathtt{|]}$
}

Default rules can be abbreviated to {\it default facts}, a new kind of atomic formulas having the general form 
$\Pi\Mathtt{\{}\Delta^{\Mathtt{+}}_1\sldots \Delta^{\Mathtt{+}}_r \  \Delta^{\Mathtt{-}}_1\sldots \Delta^{\Mathtt{-}}_s \Mathtt{\}}$, where curly braces are used instead of parentheses.
Each default fact retains an {\tt ?o}-free version of a rule's conclusion that acquires the condition predicate while omitting the condition itself.
For the special case where $r$=0 and each $\Delta^{\Mathtt{-}}_j$ is an (independent) slotted descriptor,
they correspond to ``class frame formulas''
of Flora-2~\cite{flora2-manual} \linebreak
-- when used with the compiler directive {\tt inheritance=monotonic} --
of the form
$\Pi\Mathtt{[|}\Delta^{\Mathtt{-}}_1\sldots \Delta^{\Mathtt{-}}_s \Mathtt{|]}$,
with $\Pi$ acting as the class.

For our example, the following two succinct default facts are obtained:
\begin{small}
\begin{verbatim}
Teacher{-[2 3]  % ISCO major (2: Professionals) and sub-major (3: Teaching)
        offer->service
        aptitude->explanation}

Student{acquire->KSAs  % Knowledge, Skills and Abilities
        aptitude->comprehension}
\end{verbatim}
\end{small}

PSOA's {\it default expansion} from default facts to rules is formalized by a meta-level equivalence used left to right, where $\Pi$ is an arbitrary predicate and $\Pi'$ stands for $\Pi$ [if $r$ $\geq$ 1, i.e. there is at least one dependent descriptor, $\Pi$ must be kept for its scope] or {\tt Top} [if $r$ = 0, i.e. there are no dependent descriptors,
$\Pi$ can be evacuated to {\tt Top}] ($s$ $\geq$ 0):
\begin{align*}
\Pi\Mathtt{\{} \Delta^{\Mathtt{+}}_1\sldots \Delta^{\Mathtt{+}}_r & \  \Delta^{\Mathtt{-}}_1\sldots \Delta^{\Mathtt{-}}_s \Mathtt{\}} \\
& \equiv \\
\Mathtt{Forall\ ?o\ (}
\Mathtt{?o\#}\Pi'\Mathtt{(} \Delta^{\Mathtt{+}}_1\sldots \Delta^{\Mathtt{+}}_r & \  \Delta^{\Mathtt{-}}_1\sldots \Delta^{\Mathtt{-}}_s \Mathtt{)} \ttimp{}
\Mathtt{?o\#}\Pi
\Mathtt{)}
\end{align*}

After default expansion, the inheritance querying exemplified above is realized by the PSOATransRun system
as part of its normal subpredicate-to-rule transformation and rule processing.

\section{Augmented PSOA RuleML Syntaxes}\label{sec:syntax}

Extending the abridged informal syntax introduced in Section~\ref{foundknowrep}, the unabridged formal presentation syntax and the serialization (XML) syntax of PSOA RuleML 1.03 are dealt with in this section (assuming Section~\ref{sec:inheritance}'s default facts have already been expanded to rules).
First,
the presentation syntax
is introduced in a step-wise manner, highlighting the incorporation of the dependency dimension into earlier syntaxes.
Derived from this,
the PSOA RuleML serialization syntax
is developed, focusing on how it extends atoms of Hornlog
RuleML/XML.

\subsection{PSOA RuleML Presentation Syntax}\label{sec:presentationsyntax}

We revise the syntax of \cite{Boley11d,Boley15b,ZouBoley16} to indicate the dependency dimension's dependent/independent distinction just where it is needed.
This is done such that the original syntax is reused as much as possible.

In particular, for the {\it dependent-tuple, independent-slot special case of psoa terms}, oidless or oidful, m dependent tuples and k independent slots are permitted (m $\geq$ 0, k $\geq$ 0), with tuple $i$ having length $n_i$ (1 $\leq$ $i$ $\leq$ m, $n_i$ $\geq$ 0), where a right-slot (i.e., left-tuple) normal form is assumed:
\begin{align*}
{\bf Oidless}\!:\; &\ \Mathtt{\quad f(+[t_{1,1}\sldots t_{1,n_1}]\sldots +\![t_{m,1}\sldots t_{m,n_m}]\ p_1\ttarrow{}v_1\sldots p_k\ttarrow{}v_k)} \\
{\bf Oidful}\!:\; &\ \Mathtt{o\#f(+[t_{1,1}\sldots t_{1,n_1}]\sldots +\![t_{m,1}\sldots t_{m,n_m}]\ p_1\ttarrow{}v_1\sldots p_k\ttarrow{}v_k)}
\end{align*}

We distinguish three subcases:
\begin{description}
\item[m $\geq$ 2] For {\it psoa terms with multiple dependent tuples}, ``{\tt +}''-prefixed square brackets are necessary (see above).
\item[m = 1] For {\it psoa terms with a single dependent tuple}, ``{\tt +}''-prefixed square brackets can be omitted (see {\tt 1Tupled+kSlotted} and {\tt 1Tupled} below).
\item[m = 0] For {\it tupleless psoa terms}, generalized {\it frames} arise, which as {\it framepoints} are oidful as in F-logic (see {\tt kSlotted} below) and as {\it frameships} are oidless; with
k = 0 they can additionally be specialized to {\it slotless psoa terms}, arriving at
{\it empty psoa terms}, for which round parentheses can be omitted in the oidful case (see {\tt Membership} below).
\end{description}

Starting with the below oidful psoa terms, color-coding shows syntactic variants for the subcases m = 1 and m = k = 0 (\textcolor{magenta}{single-dependent-tuple brackets} for $n\tsub{1}$ $\geq$ 1 are optional, as are
\textcolor{blue}{zero-argument parentheses}):
\\[0.8em]
{\tt
\begin{tabular}{@{}ll@{\:}l@{}l}
1Tupled+kSlotted: & o\hash f(\textcolor{magenta}{+[}t\tsub{1,1}\:...\:t\tsub{1,n\tsub{1}}\textcolor{magenta}{]} & p\tsub{1}->v\tsub{1}\:...\:p\tsub{k}->v\tsub{k}) \\
1Tupled: & o\hash f(\textcolor{magenta}{+[}t\tsub{1,1}\:...\:t\tsub{1,n\tsub{1}}\textcolor{magenta}{]}) & \\
kSlotted: & o\hash f( & p\tsub{1}->v\tsub{1}\:...\:p\tsub{k}->v\tsub{k}) \\
Membership: &  o\hash f\textcolor{blue}{()} \\
\end{tabular}
}
\vspace{0.4em}

\pagebreak  
Moving on to the {\it dependent/independent-tuple, dependent/independent-slot general case of oidful psoa terms}, below we obtain four
subsequences for the four bags of descriptor variety and dependency (in the pretty-print arranged as four separate lines). Here, the superscripts indicate subterms that are part of dependent (``$^{\Mathtt{+}}$'') 
vs. independent (``$^{-}$'')
descriptors. Refining earlier PSOA versions, a {\it right-slot, right-independent} (i.e., left-tuple, left-dependent) normal form is assumed.
As suggested by the order of the italicized qualifiers, this normal form first distinguishes the descriptor variety and second the descriptor dependency ($m^{+}$ $\geq$ 0, $m^{-}$ $\geq$ 0, $k^{+}$ $\geq$ 0, $k^{-}$ $\geq$ 0, and for 1 $\leq$ $i^{+/-}$ $\leq$ $m^{+/-}$, $n^{+}_{i^{+}}$ $\geq$ 0, $n^{-}_{i^{-}}$ $\geq$ 0):
\begin{align*}
\Mathtt{o\#f(} & \Mathtt{+[t^{+}_{1,1}\sldots t^{+}_{1,n^{+}_1}]\sldots +\![t^{+}_{m^{+},1}\sldots t^{+}_{m^{+},n^{+}_{m^{+}}}]} \\
& \Mathtt{-[t^{-}_{1,1}\sldots t^{-}_{1,n^{-}_1}]\sldots -\![t^{-}_{m^{-},1}\sldots t^{-}_{m^{-},n^{-}_{m^{-}}}]} \\
& \Mathtt{p^{+}_1\ttarrowdep{}v^{+}_1\sldots p^{+}_{k^{+}}\ttarrowdep{}v^{+}_{k^{+}}} \\
& \Mathtt{p^{-}_1\ttarrow{}v^{-}_1\sldots p^{-}_{k^{-}}\ttarrow{}v^{-}_{k^{-}})}
\end{align*}

For formulating the laws in Section~\ref{lawsknowtrans} using the abstract-descriptor-level \linebreak
pattern
$\Delta^{\Mathtt{+}}_1\sldots \Delta^{\Mathtt{+}}_r \  \Delta^{\Mathtt{-}}_1\sldots \Delta^{\Mathtt{-}}_s$,
an equivalent {\it right-independent} (i.e., left-dependent) form was assumed for convenience, which
could be refined to an equivalent \linebreak
{\it right-independent, right-slot} (i.e., left-dependent, left-tuple) form:
\begin{align*}
\Mathtt{o\#f(} & \Mathtt{+[t^{+}_{1,1}\sldots t^{+}_{1,n^{+}_1}]\sldots +\![t^{+}_{m^{+},1}\sldots t^{+}_{m^{+},n^{+}_{m^{+}}}]} \\
& \Mathtt{p^{+}_1\ttarrowdep{}v^{+}_1\sldots p^{+}_{k^{+}}\ttarrowdep{}v^{+}_{k^{+}}} \\
& \Mathtt{-[t^{-}_{1,1}\sldots t^{-}_{1,n^{-}_1}]\sldots -\![t^{-}_{m^{-},1}\sldots t^{-}_{m^{-},n^{-}_{m^{-}}}]} \\
& \Mathtt{p^{-}_1\ttarrow{}v^{-}_1\sldots p^{-}_{k^{-}}\ttarrow{}v^{-}_{k^{-}})}
\end{align*}


The below EBNF grammar for the presentation syntax since PSOA RuleML 1.0 uses
a right-slot form (all slots are to the right of all tuples) but not any dependency form (the order between dependent vs. independent descriptors is not prescribed). It advances the grammar of the earlier PSOA RuleML \cite{Boley11d} as follows:
\begin{itemize}
\item
Employs the document root {\tt RuleML}, rather than the earlier {\tt Document}, as well as {\tt Assert}, rather than the earlier {\tt Group}, complementing it with {\tt Query}, which was absent earlier.
\item
Refines both varieties of descriptors for the (``{\tt DI}''-)distinction of Dependent vs. Independent tuples ({\tt TUPLEDI}) and slots ({\tt SLOTDI}).
\item
Reflects the use of
(a) oidless and oidful psoa terms as {\tt Atom}s in/as {\tt FORMULA}s,
(b) oidful {\tt Atom}s (for unnesting, leaving behind the OID term) as {\tt TERM}s in {\tt Atom}s and {\tt Expr}essions,
as well as
(c) oidless psoa terms as {\tt Expr}essions.
\item Revises the {\tt CLAUSE}, {\tt Implies}, and {\tt HEAD} productions to make the PSOA RuleML language closed under objectification and describution.
\end{itemize}

On the top-level, the EBNF grammar is divided into two parts: Basically, while the {\it Rule Language} provides ``wrapper'' declarations around rules and the upper levels of the rules themselves, the {\it Condition Language} provides the formula specification for the rule conditions and for queries, and also defines psoa terms.
\pagebreak   
\\[1.0em]
\textbf{Rule Language:}
\begin{small}
\begin{verbatim}
RuleML ::= 'RuleML' '(' Base? Prefix* Import* (Assert | Query)* ')'
Base ::= 'Base' '(' ANGLEBRACKIRI ')'
Prefix ::= 'Prefix' '(' Name ANGLEBRACKIRI ')'
Import ::= 'Import' '(' ANGLEBRACKIRI PROFILE? ')'
Assert ::= 'Assert' '(' (RULE | Assert)* ')'
Query ::= 'Query' '(' FORMULA ')'
RULE ::= ('Forall' Var+ '(' CLAUSE ')') | CLAUSE
CLAUSE ::= Implies | HEAD
Implies ::= HEAD ':-' FORMULA
HEAD ::= ATOMIC | 'Exists' Var+ '(' HEAD ')' | 'And' '(' HEAD* ')'
PROFILE ::= ANGLEBRACKIRI
\end{verbatim}
\end{small}
\vspace{0.8em}
\textbf{Condition Language:}\footnote{Constant/variable names since PSOA 1.0 use a simplified
SPARQL 1.1 production [169] for {\tt PN\_LOCAL} (\url{https://www.w3.org/TR/sparql11-query/\#sparqlGrammar}),
with sources on GitHub (\url{https://github.com/RuleML/PSOATransRunComponents/blob/master/PSOACore/src/main/antlr3/org/ruleml/psoa/parser/PSOAPS.g}).
In this ANTLR grammar, the {\tt CONSTSHORT}-produced (rather than {\tt Var}-produced) occurrence of {\tt PN\_LOCAL} 
is further differentiated -- by an embedded Java action -- 
into
an ``{\tt \_}''-prefixed or unprefixed local name
(a stand-alone ``{\tt \_}'' works separately).

}
\begin{small}
\begin{verbatim}
FORMULA ::= 'And' '(' FORMULA* ')' |    % Main start symbol: formulas
            'Or' '(' FORMULA* ')' |
            'Exists' Var+ '(' FORMULA ')' |
            ATOMIC |
            'External' '(' Atom ')'
ATOMIC ::= Atom | Equal | Subclass
Atom ::= ATOMOIDLESS | ATOMOIDFUL  % Atoms can be oidless or oidful
ATOMOIDLESS ::= PSOAOIDLESS        % Oidless atoms are oidless psoa terms
ATOMOIDFUL ::= PSOAOIDFUL          % Oidful atoms are oidful psoa terms
Equal ::= TERM '=' TERM
Subclass ::= TERM '##' TERM  % Subclass is pars pro toto for Subpredicate
PSOA ::= PSOAOIDLESS | PSOAOIDFUL       % Extra start symbol: psoa terms
PSOAOIDLESS ::= TERM '(' (TERM* | TUPLEDI*) SLOTDI* ')'
PSOAOIDFUL ::= TERM '#' PSOAOIDLESS
TUPLEDI ::= ('+' | '-') '[' TERM* ']'
SLOTDI ::= TERM ('+>' | '->') TERM
TERM ::= Const | Var | ATOMOIDFUL | Expr | 'External' '(' Expr ')'
Expr ::= PSOAOIDLESS               % Exprs are oidless psoa terms
Const ::= '"' UNICODESTRING '"^^' SYMSPACE | CONSTSHORT
Var ::= '?' PN_LOCAL?
SYMSPACE ::= ANGLEBRACKIRI | CURIE
CONSTSHORT ::= ANGLEBRACKIRI | CURIE | '"' UNICODESTRING '"'
                                              | NumericLiteral | PN_LOCAL
\end{verbatim}
\end{small}

\mycomment{
\begin{verbatim}
    FORMULA ::= 'And' '(' FORMULA* ')' |
                'Or' '(' FORMULA* ')' |
                'Exists' Var+ '(' FORMULA ')' |
                ATOMIC |
                'External' '(' Atom ')'
    ATOMIC ::= Atom | Equal | Subclass

    Equal ::= TERM '=' TERM
    Subclass ::= TERM '##' TERM  % Subclass is pars pro toto for Subpredicate
    Atom    ::= (TERM '#')? TERM '(' (TERM* | TUPLEDI*) SLOTDI* ')'


    Expr    ::= TERM '(' (TERM* | TUPLEDI*) SLOTDI* ')'


    TUPLEDI ::= ('+' | '-') '[' TERM* ']'
    SLOTDI  ::= TERM ('+>' | '->') TERM
    TERM ::= Const | Var | Atom | Expr | 'External' '(' Expr ')'
    Const ::= '"' UNICODESTRING '"^^' SYMSPACE | CONSTSHORT
    Var ::= '?' UNICODESTRING?
    SYMSPACE ::= ANGLEBRACKIRI | CURIE
\end{verbatim}

\begin{verbatim}
    PSOA ::= Atom | EXPR
    Atom ::= ATOMOIDLESS | ATOMOIDFUL
    ATOMOIDLESS ::= TERM '(' (TERM* | TUPLEDI*) SLOTDI* ')'
    ATOMOIDFUL ::= TERM '#' ATOMOIDLESS
    
    TERM ::= Const | Var | ATOMOIDFUL | Expr | 'External' '(' Expr ')'
    Expr ::= TERM '(' (TERM* | TUPLEDI*) SLOTDI* ')'
\end{verbatim}

\begin{verbatim}
    Atom ::= PSOA
    PSOA ::= PSOAOIDLESS | PSOAOIDFUL
    PSOAOIDLESS ::= TERM '(' (TERM* | TUPLEDI*) SLOTDI* ')'
    PSOAOIDFUL ::= TERM '#' PSOAOIDLESS
    TERM ::= Const | Var | PSOA | 'External' '(' Expr ')'
    Expr ::= PSOAOIDLESS
\end{verbatim}
}

Examples for KBs according to a wrapperless Rule Language and the \linebreak
Condition Language were given
in Sections \ref{foundknowrep} and \ref{lawsknowtrans}.


Fig.~\ref{SampleKB} shows a sample KB, called {\tt (KB)}, for the {\tt RuleML}/{\tt Assert}-wrapped
Rule Language adding {\tt (R1)} to the Condition Language clauses of the correspondingly modified {\tt (KB2)}, called {\tt (KB2\#)}.
Note that the {\tt RuleML} wrapper contains a {\tt Prefix} statement for defining, CURIE-like, {\tt pred:} to access W3C RIF built-in predicates~\cite{PolleresBoley+13} from within the {\tt Assert}-wrapped PSOA KB.
Fig.~\ref{SampleKB} also includes
the optional, hence again (blue-)colored, ``\textcolor{blue}{{\tt \_}}'' prefix for local constants,
and can be copied \& pasted in any ``\textcolor{blue}{{\tt \_}}'' 
form into a {\tt *.psoa} KB file for PSOATransRun.

\begin{figure*}
\begin{small}
\begin{verbatim}
RuleML (
  Prefix(pred: <http://www.w3.org/2007/rif-builtin-predicate#>)
\end{verbatim}

\begin{Verbatim}[commandchars=\\\{\}]  
  Assert (                      % (KB)
                                   % (KB2#)
    \textcolor{blue}{_}Teacher##\textcolor{blue}{_}Scholar               % Taxonomy
    \textcolor{blue}{_}Student##\textcolor{blue}{_}Scholar
    \textcolor{blue}{_}TA##\textcolor{blue}{_}Teacher
    \textcolor{blue}{_}TA##\textcolor{blue}{_}Student
                                     % Data (ii)
    \textcolor{blue}{_}John#\textcolor{blue}{_}Teacher(+[\textcolor{blue}{_}Wed \textcolor{blue}{_}Thu]
                   \textcolor{blue}{_}coursehours+>12 \textcolor{blue}{_}dept+>\textcolor{blue}{_}Physics \textcolor{blue}{_}salary+>29400
                                                    \textcolor{blue}{_}income->29400)
    \textcolor{blue}{_}John#\textcolor{blue}{_}Student(+[\textcolor{blue}{_}Mon \textcolor{blue}{_}Tue \textcolor{blue}{_}Fri] -[1995 8 17] 
                   \textcolor{blue}{_}coursehours+>20 \textcolor{blue}{_}dept+>\textcolor{blue}{_}Math \textcolor{blue}{_}gender->\textcolor{blue}{_}male)
    
    Forall ?o ?ht ?hs (            % (R1)
      ?o#\textcolor{blue}{_}TA(\textcolor{blue}{_}workload+>\textcolor{blue}{_}high) :-
        And(?o#\textcolor{blue}{_}Teacher(\textcolor{blue}{_}coursehours+>?ht)
            External(pred:numeric-greater-than(?ht 10))    % ?ht > 10
            ?o#\textcolor{blue}{_}Student(\textcolor{blue}{_}coursehours+>?hs)
            External(pred:numeric-greater-than(?hs 18)))   % ?hs > 18
    )
  )
)
\end{Verbatim}
\end{small}
\vspace{-0.5cm}
\caption{Sample {\tt (KB)} of taxonomy plus data, constituting {\tt (KB2\#)}, and rule {\tt (R1)}
in \textcolor{blue}{un}abridged presentation syntax for PSOATransRun translation \& execution.}
\label{SampleKB}
\end{figure*}


\vspace{-0.5em}
\subsection{PSOA RuleML Serialization Syntax}\label{sec:serializationsyntax}

The PSOA RuleML/XML 1.03 serialization syntax extends the one of Hornlog RuleML/XML 1.03.
The XML serialization syntax of PSOA RuleML 1.03 can be derived from the presentation syntax.
Besides obvious differences due to its use of XML markup, the serialization syntax mainly differs from the presentation syntax in being ``striped'', alternating between -- (Java-method-style) all-lower-cased -- {\it edges} (absent from the presentation syntax) and  -- (Java-class-style)
first-letter-upper-cased -- {\it Nodes} (having counterparts in the presentation syntax).
While edges and Nodes are non-terminals that are `visible' in the (parsed or
generated) serialization syntax, there are also traditional -- all-upper-cased -- non-terminals that are `invisible'.\footnote{We employ Relax NG as the main language to define schemas for XML, where \linebreak
`visible' non-terminals correspond to {\tt element} patterns and
`invisible' non-terminals correspond to {\it named patterns} (\url{http://relaxng.org/compact-tutorial-20030326.html\#id2814516}).}

For the core (dependent and independent) descriptor-defining EBNF-grammar productions of the presentation syntax in Section \ref{sec:presentationsyntax} (reproduced -- slightly modified -- with a ``P(resentation):'' label), we give below corresponding
EBNF-like productions for the serialization syntax  (introduced with an ``X(ML):'' label).
\\[1.0em]
\textbf{Condition Language Descriptors (Presentation to Serialization):}
\begin{small}
\begin{verbatim}
P:  TUPLEDI ::= '+' '[' TERM* ']' | '-' '[' TERM* ']'
X:  TUPLEDI ::= tupdep | tup   % Different edges
X:  tupdep ::= Tuple           % lead into same
X:  tup ::= Tuple              % Tuple Node

P:  SLOTDI ::= TERM '+>' TERM | TERM '->' TERM
X:  SLOTDI ::= slotdep | slot  % Different edges
X:  slotdep ::= TERM TERM      % lead into same
X:  slot ::= TERM TERM         % pair of TERM Nodes
\end{verbatim}
\end{small}

Entire atoms with such in/dependent-tuple, 
in/dependent-slot descriptors in serialization syntax can be similarly derived from the presentation syntax of Section \ref{sec:presentationsyntax}.
This can be used to parse or generate XML-serialized atoms as follows:
\begin{small}
\begin{lstlisting}[mathescape]
<Atom>
  <oid><Ind>$\Mathtt{o}$</Ind></oid><op><Rel>$\Mathtt{f}$</Rel></op>
  <tupdep><Tuple>$\Mathtt{t^{+}_{1,1}\sldots t^{+}_{1,n^{+}_1}}$</Tuple></tupdep>$\ .\ .\ .\ $
       <tupdep><Tuple>$\Mathtt{t^{+}_{m^{+},1}\sldots t^{+}_{m^{+},n^{+}_{m^{+}}}}$</Tuple></tupdep>
  <tup><Tuple>$\Mathtt{t^{-}_{1,1}\sldots t^{-}_{1,n^{-}_1}}$</Tuple></tup>$\ .\ .\ .\ $
       <tup><Tuple>$\Mathtt{t^{-}_{m^{-},1}\sldots t^{-}_{m^{-},n^{-}_{m^{-}}}}$</Tuple></tup>
  <slotdep>$\Mathtt{p^{+}_1 \; v^{+}_1}$</slotdep>$\ .\ .\ .\ $<slotdep>$\Mathtt{p^{+}_{k^{+}} \; v^{+}_{k^{+}}}$</slotdep>
  <slot>$\Mathtt{p^{-}_1 \; v^{-}_1}$</slot>$\ .\ .\ .\ $<slot>$\Mathtt{p^{-}_{k^{-}} \; v^{-}_{k^{-}}}$</slot>
</Atom>
\end{lstlisting}
\end{small}
Here, the meta-variables $\Mathtt{o}$ and $\Mathtt{f}$ as well as the decorated meta-variables $\Mathtt{t}$, $\Mathtt{p}$, and $\Mathtt{v}$ indicate, respectively, recursively
XML-serialized OIDs and predicates \linebreak
as well as terms, properties, i.e. slot names, and values, i.e. slot fillers,
of their presentation-syntax versions in Section~\ref{sec:presentationsyntax}.

The three psoa-atom ground facts of Section \ref{factsandqueries}'s Rich TA example {\tt (KB2)} in presentation syntax result in the following serialization syntax  (color-coded as in Fig.~\ref{TAexample} and {\tt (KB2)}):
\mycomment{
\begin{small}
\begin{verbatim}
<Atom>
  <oid><Ind>John</Ind></oid><op><Rel>TA</Rel></op>
  <slotdep><Ind>workload</Ind><Ind>high</Ind></slotdep>
</Atom>

<Atom>
  <oid><Ind>John</Ind></oid><op><Rel>Teacher</Rel></op>
  <tupdep><Tuple><Ind>Wed</Ind><Ind>Thu</Ind></Tuple></tupdep>
  <slotdep><Ind>dept</Ind><Ind>Physics</Ind></slotdep>
  <slotdep><Ind>salary</Ind><Ind>29400</Ind></slotdep>
  <slot><Ind>income</Ind><Ind>29400</Ind></slot>
</Atom>

<Atom>
  <oid><Ind>John</Ind></oid><op><Rel>Student</Rel></op>
  <tupdep>
    <Tuple><Ind>Mon</Ind><Ind>Tue</Ind><Ind>Fri</Ind></Tuple>
  </tupdep>
  <tup>
    <Tuple><Ind>1995</Ind><Ind>8</Ind><Ind>17</Ind></Tuple>
  </tup>
  <slotdep><Ind>dept</Ind><Ind>Math</Ind></slotdep>
  <slot><Ind>gender</Ind><Ind>male</Ind></slot>
</Atom>
\end{verbatim}
\end{small}
}
{
\color{brown}
\begin{small}
\begin{Verbatim}[commandchars=\\\{\}]
{\bf\scriptsize<Atom>}
  {\bf\scriptsize<oid><Ind>}John{\bf\scriptsize</Ind></oid><op><Rel>}TA{\bf\scriptsize</Rel></op>}
  {\bf\scriptsize<slotdep><Ind>}workload{\bf\scriptsize</Ind><Ind>}high{\bf\scriptsize</Ind></slotdep>}
{\bf\scriptsize</Atom>}
\end{Verbatim}
\end{small}
}
\vspace{-1.2em}
{
\color{red}
\begin{small}
\begin{Verbatim}[commandchars=\\\{\}]
{\bf\scriptsize<Atom>}
  {\bf\scriptsize<oid><Ind>}John{\bf\scriptsize</Ind></oid><op><Rel>}Teacher{\bf\scriptsize</Rel></op>}
  {\bf\scriptsize<tupdep><Tuple><Ind>}Wed{\bf\scriptsize</Ind><Ind>}Thu{\bf\scriptsize</Ind></Tuple></tupdep>}
  {\bf\scriptsize<slotdep><Ind>}dept{\bf\scriptsize</Ind><Ind>}Physics{\bf\scriptsize</Ind></slotdep>}
  {\bf\scriptsize<slotdep><Ind>}salary{\bf\scriptsize</Ind><Ind>}29400{\bf\scriptsize</Ind></slotdep>}
  {\bf\scriptsize<slot><Ind>}income{\bf\scriptsize</Ind><Ind>}29400{\bf\scriptsize</Ind></slot>}
{\bf\scriptsize</Atom>}
\end{Verbatim}
\end{small}
}
\vspace{-1.2em}
{
\color{ForestGreen}
\begin{small}
\begin{Verbatim}[commandchars=\\\{\}]
{\bf\scriptsize<Atom>}
  {\bf\scriptsize<oid><Ind>}John{\bf\scriptsize</Ind></oid><op><Rel>}Student{\bf\scriptsize</Rel></op>}
  {\bf\scriptsize<tupdep>}
    {\bf\scriptsize<Tuple><Ind>}Mon{\bf\scriptsize</Ind><Ind>}Tue{\bf\scriptsize</Ind><Ind>}Fri{\bf\scriptsize</Ind></Tuple>}
  {\bf\scriptsize</tupdep>}
  {\bf\scriptsize<tup>}
    {\bf\scriptsize<Tuple><Ind>}1995{\bf\scriptsize</Ind><Ind>}8{\bf\scriptsize</Ind><Ind>}17{\bf\scriptsize</Ind></Tuple>}
  {\bf\scriptsize</tup>}
  {\bf\scriptsize<slotdep><Ind>}dept{\bf\scriptsize</Ind><Ind>}Math{\bf\scriptsize</Ind></slotdep>}
  {\bf\scriptsize<slot><Ind>}gender{\bf\scriptsize</Ind><Ind>}male{\bf\scriptsize</Ind></slot>}
{\bf\scriptsize</Atom>}
\end{Verbatim}
\end{small}
}


The serialization for the rest of the Condition Language and for the Rule Language can be derived analogously. 

A Relax NG schema for PSOA RuleML 1.03 has been developed, first in a monolithic manner\footnote{\url{http://wiki.ruleml.org/index.php/PSOA_RuleML\#Monolithic_Syntax}} and then using Relax NG's modularization capability\footnote{\url{http://wiki.ruleml.org/index.php/PSOA_RuleML\#Modular_Syntax}}.

\section{Model-Theoretic PSOA RuleML Semantics}\label{sec:semantics}
\newcommand\setOfBags[1]{\functt{SetOfFiniteBags}{#1}}
\newcommand\setOIDEmpty{\functt{SetOfPhiSingletons}{\dind{}}}

Key parts of the formal model-theoretic semantics definitions from \cite{Boley11d,ZouBoley16} are revised here for PSOA RuleML 1.0's and 1.03's object-virtualizing, in/dependent-tuple, in/dependent-slot psoa terms in
right-slot, right-independent normal form (cf. Section~\ref{sec:presentationsyntax}).
The revised definitions will be evolved from the earlier ones.

Truth valuation of PSOA RuleML formulas is defined as a mapping \textit{TVal}\calsubscript{I} in two steps:
1. A mapping \textbf{\textit{I}} generically bundles various mappings from a \textbf{\textit{semantic structure}}, $\cal{I}$; \textbf{\textit{I}} maps a formula to an element of the domain \textbf{\textit{D}}. \linebreak
2. A mapping \textbf{\textit{I}}\textsubscript{truth}
takes such a \textbf{\textit{D}} element to a set of \textbf{\textit{truth values}}, \textbf{\textit{TV}}, here fixed to the set $\{\textbf{t},\textbf{f}\}$ of classical two-valued logic. 
For the interpretation of individuals, \textbf{\textit{D}}\textsubscript{ind}, a non-empty subset of \textbf{\textit{D}}, is used.

As a central part of
$\cal{I}$,
Definition 4, case 3, of \cite{Boley11d}
introduced the total mapping \textbf{\textit{I}}\textsubscript{psoa} without yet specifying object virtualization, of \cite{ZouBoley16}, nor dependency:
\linebreak
\textbf{\textit{I}}\textsubscript{psoa}
mapped
\textbf{\textit{D}} to total functions that have the general 3-ary form \linebreak
\textbf{\textit{D}}\textsubscript{ind}
{\texttimes}
\texttt{SetOfFiniteBags}(\textbf{\textit{D*}}\textsubscript{ind})
{\texttimes}
\texttt{SetOfFiniteBags}(\textbf{\textit{D}}\textsubscript{ind}
{\texttimes} \textbf{\textit{D}}\textsubscript{ind}) ${\rightarrow}$
\textbf{\textit{D}}. \linebreak
An argument \texttt{d}
${\in}$ \textbf{\textit{D}} of
\textbf{\textit{I}}\textsubscript{psoa} uniformly represents the
function or predicate 
symbol \texttt{f} of psoa terms \texttt{o\#f(\ldots)}.
An element \texttt{c} ${\in}$ \textbf{\textit{D}}\textsubscript{ind}
in the first Cartesian argument 
of the resulting total functions
represents an object
as the interpretation of \texttt{o} from \texttt{o\#f},
where \texttt{c} is described with two bags
in the second and third Cartesian arguments 
(enclosed by ``\{\ldots\}'', but allowing repeated elements):
\begin{itemize}
\item
A finite bag of finite tuples \{{\textless}\texttt{t}\texttt{\textsubscript{1,1}},
..., \texttt{t\textsubscript{1,n\textsubscript{1}}}{\textgreater}, \texttt{...},
{\textless}\texttt{t}\texttt{\textsubscript{m,1}},
..., \texttt{t\textsubscript{m,n\textsubscript{m}}}{\textgreater}\} ${\in}$ \texttt{SetOfFiniteBags}(\textbf{\textit{D*}}\textsubscript{ind}), possibly empty,
represents positional information.
Here,
\textbf{\textit{D*}}\textsubscript{ind} is the
set of all finite tuples over the domain
\textbf{\textit{D}}\textsubscript{ind}.
\item
A finite bag of attribute-value pairs
\{{\textless}\texttt{a}\texttt{\textsubscript{1}},\texttt{v}\texttt{\textsubscript{1}}{\textgreater}, \texttt{...},
{\textless}\texttt{a}\texttt{\textsubscript{k}},\texttt{v}\texttt{\textsubscript{k}}{\textgreater}\}
${\in}$ \linebreak
\texttt{SetOfFiniteBags}(\textbf{\textit{D}}\textsubscript{ind}
{\texttimes} \textbf{\textit{D}}\textsubscript{ind}), possibly empty,
for slotted information.
\end{itemize}

For PSOA RuleML 1.0 and 1.03 (henceforth: 1.0\&1.03), the definition of \ipsoa{} is revised to map \dom{} to total functions of the general 5-ary form 
\vspace{-0.6em}
\begin{multline*}
\ \ \ \ \ \ \setOIDEmpty \\
\times\setOfBags{\dind[*]} 
\times\setOfBags{\dind[*]} \ \ \ \ \ \ \ \ \ \ \ \ \ \ \ \ \\
\times\setOfBags{\dind{}\times\dind{}} 
\times\setOfBags{\dind{}\times\dind{}} \\
\rightarrow\dom{}
\end{multline*}
where
the argument in the first line interprets the possibly virtualized object,
the two arguments of the same type in the second line interpret dependent and independent tuples, and
the two arguments of the same type in the third line interpret dependent and independent slots
(thus
the earlier two bags are refined to four).
Also,
\setOIDEmpty{}, from \cite{ZouBoley16},
is defined as $\{\{\}\}\cup\{\{{\tt c}\}\mid\Mathtt{c\in\dind}\}$, whose elements are the empty set \{\} and a singleton set $\{\Mathtt{c}\}$ for each $\Mathtt{c\in\dind}$. With this definition, the total function resulting from \ipsoa{\mathrm{\ipmap{f}}} can be appropriately applied to its arguments in Equations~(\ref{def:oidful-psoa-ip}) and~(\ref{def:oidless-psoa-ip}) below.

The generic recursive mapping \textbf{\textit{I}} is defined from terms to their subterms and ultimately to \textbf{\textit{D}}.
In \cite{Boley11d}, Definition 4 
-- before the differentiation of in/dependent descriptors -- the mapping of psoa terms was as follows:
\\[0.2em]
\textbf{\textit{I}}(\texttt{o\#f\openps{}}\texttt{[t}\texttt{\textsubscript{1,1}}
... \texttt{t\textsubscript{1,n\textsubscript{1}}}\texttt{]} \texttt{...}
\texttt{[t}\texttt{\textsubscript{m,1}}
... \texttt{t\textsubscript{m,n\textsubscript{m}}}\texttt{]} \texttt{a}\texttt{\textsubscript{1}}\texttt{->v}\texttt{\textsubscript{1}}\texttt{
...
a}\texttt{\textsubscript{k}}\texttt{->v}\texttt{\textsubscript{k}}\texttt{\closeps{}})
= \newline
\textbf{\textit{I}}\textsubscript{psoa}(\textbf{\textit{I}}(\texttt{f}))(\textbf{\textit{I}}(\texttt{o}),  \\
  \hspace*{1.65cm}
\{{\textless}\textbf{\textit{I}}(\texttt{t}\texttt{\textsubscript{1,1}}),
..., \textbf{\textit{I}}(\texttt{t\textsubscript{1,n\textsubscript{1}}}){\textgreater},
...,
{\textless}\textbf{\textit{I}}(\texttt{t}\texttt{\textsubscript{m,1}}),
..., \textbf{\textit{I}}(\texttt{t\textsubscript{m,n\textsubscript{m}}}){\textgreater}\},  \\
  \hspace*{1.65cm}
\{{\textless}\textbf{\textit{I}}(\texttt{a}\texttt{\textsubscript{1}}),\textbf{\textit{I}}(\texttt{v}\texttt{\textsubscript{1}}){\textgreater},
...,
{\textless}\textbf{\textit{I}}(\texttt{a}\texttt{\textsubscript{k}}),\textbf{\textit{I}}(\texttt{v}\texttt{\textsubscript{k}}){\textgreater}\})
\\[0.3em]
In PSOA RuleML 1.0\&1.03,
for {\bf oidful} psoa terms, the definition of \ipmap{}
becomes:
\vspace{-1.8em}
\begin{multline}\label{def:oidful-psoa-ip}
\ipmap{
\begin{array}{r@{}l}
\Mathtt{o\#f(} & \Mathtt{+[t^{+}_{1,1}\sldots t^{+}_{1,n^{+}_1}]\sldots +\![t^{+}_{m^{+},1}\sldots t^{+}_{m^{+},n^{+}_{m^{+}}}]} \\
& \Mathtt{-[t^{-}_{1,1}\sldots t^{-}_{1,n^{-}_1}]\sldots -\![t^{-}_{m^{-},1}\sldots t^{-}_{m^{-},n^{-}_{m^{-}}}]} \\
& \Mathtt{p^{+}_1\ttarrowdep{}v^{+}_1\sldots p^{+}_{k^{+}}\ttarrowdep{}v^{+}_{k^{+}}} \\
& \Mathtt{p^{-}_1\ttarrow{}v^{-}_1\sldots p^{-}_{k^{-}}\ttarrow{}v^{-}_{k^{-}})}
\end{array}
} = \\
\begin{aligned}
\ipsoa{\mathrm{\ipmap{f}}}(& \{\ipmap{o}\}, \\
              & \{\angles{\ipmap{t^{+}_{1,1}},\ldots,\ipmap{t^{+}_{1,n^{+}_1}}},\ldots,\angles{\ipmap{t^{+}_{m^{+},1}},\ldots,\ipmap{t^{+}_{m^{+},n^{+}_{m^{+}}}}}\}, \\
              & \{\angles{\ipmap{t^{-}_{1,1}},\ldots,\ipmap{t^{-}_{1,n^{-}_1}}},\ldots,\angles{\ipmap{t^{-}_{m^{-},1}},\ldots,\ipmap{t^{-}_{m^{-},n^{-}_{m^{-}}}}}\}, \\
              & \{\angles{\ipmap{p^{+}_1},\ipmap{v^{+}_1}},\ldots,\angles{\ipmap{p^{+}_{k^{+}}},\ipmap{v^{+}_{k^{+}}}}\} \\
              & \{\angles{\ipmap{p^{-}_1},\ipmap{v^{-}_1}},\ldots,\angles{\ipmap{p^{-}_{k^{-}}},\ipmap{v^{-}_{k^{-}}}}\})
\end{aligned}
\end{multline}
\normalsize
Here, the first argument of the semantic function \ipsoa{\mathrm{\ipmap{f}}} is wrapped into a singleton set $\{\ipmap{o}\}$ \cite{ZouBoley16}, the second and third arguments are
interpretations of, respectively, dependent and independent tuples, and the fourth and fifth arguments are
interpretations of, respectively, dependent and independent slots. 
The first-argument wrapping method in Equation~(\ref{def:oidful-psoa-ip}) specializes to using the empty set \{\} as the first argument
to separately define
\ipmap{} for {\bf oidless} psoa terms:
\begin{multline}\label{def:oidless-psoa-ip}
\ipmap{
\begin{array}{r@{}l}
\Mathtt{f(} & \Mathtt{+[t^{+}_{1,1}\sldots t^{+}_{1,n^{+}_1}]\sldots +\![t^{+}_{m^{+},1}\sldots t^{+}_{m^{+},n^{+}_{m^{+}}}]} \\
& \Mathtt{-[t^{-}_{1,1}\sldots t^{-}_{1,n^{-}_1}]\sldots -\![t^{-}_{m^{-},1}\sldots t^{-}_{m^{-},n^{-}_{m^{-}}}]} \\
& \Mathtt{p^{+}_1\ttarrowdep{}v^{+}_1\sldots p^{+}_{k^{+}}\ttarrowdep{}v^{+}_{k^{+}}} \\
& \Mathtt{p^{-}_1\ttarrow{}v^{-}_1\sldots p^{-}_{k^{-}}\ttarrow{}v^{-}_{k^{-}})}
\end{array}
} = \\
\begin{aligned}
\ipsoa{\mathrm{\ipmap{f}}}(& \{\}, \\
              & \{\angles{\ipmap{t^{+}_{1,1}},\ldots,\ipmap{t^{+}_{1,n^{+}_1}}},\ldots,\angles{\ipmap{t^{+}_{m^{+},1}},\ldots,\ipmap{t^{+}_{m^{+},n^{+}_{m^{+}}}}}\}, \\
              & \{\angles{\ipmap{t^{-}_{1,1}},\ldots,\ipmap{t^{-}_{1,n^{-}_1}}},\ldots,\angles{\ipmap{t^{-}_{m^{-},1}},\ldots,\ipmap{t^{-}_{m^{-},n^{-}_{m^{-}}}}}\}, \\
              & \{\angles{\ipmap{p^{+}_1},\ipmap{v^{+}_1}},\ldots,\angles{\ipmap{p^{+}_{k^{+}}},\ipmap{v^{+}_{k^{+}}}}\} \\
              & \{\angles{\ipmap{p^{-}_1},\ipmap{v^{-}_1}},\ldots,\angles{\ipmap{p^{-}_{k^{-}}},\ipmap{v^{-}_{k^{-}}}}\})
\end{aligned}
\end{multline}

When, as in the below Definition 5, case 3, \textbf{\textit{I}} is applied to a psoa term, its total function is obtained from \textbf{\textit{I}}\textsubscript{psoa} applied to the recursively interpreted predicate argument \texttt{f}.
The application of the resulting total function to the recursively interpreted other parts of a psoa term denotes the term's interpretation in \textbf{\textit{D}}.
Because PSOA RuleML's model theory has incorporated oidless psoa terms since \cite{ZouBoley16}, as reflected by the above Equation~(\ref{def:oidless-psoa-ip}), it could not uniformly use the (interpreted) OID \texttt{o} as the \textbf{\textit{I}}\textsubscript{psoa} argument.
Instead, already since \cite{Boley11d}, it has uniformly used the (interpreted) predicate \texttt{f},
which is justified by the predicate \texttt{f} always being present and user-controlled for psoa terms,
with increasing precision when descending the taxonomy from
the `catch-all' total function obtained from \textbf{\textit{I}}\textsubscript{psoa} applied to the interpretation $\top$ of the root predicate \texttt{Top}.
On the other hand, the OID \texttt{o} -- which in RIF-BLD is used for the \textbf{\textit{I}}\textsubscript{frame} argument -- need not be user-controlled in PSOA but can be system-generated via objectification, e.g. as an existential variable or a (Skolem) constant, so is not suited to obtain a meaningful total function for a psoa term.
When applied to the same predicate used in different psoa terms, \textbf{\textit{I}}\textsubscript{psoa} obtains the same total function, which when itself applied to different psoa terms can return the same or different values.

In PSOA RuleML 1.0\&1.03, the earlier \cite{Boley11d} Definition 5, case 3, is revised by recursively defining truth valuation \textit{TVal}\calsubscript{I} for psoa formulas, based on the above-revised \textbf{\textit{I}} and on the mapping \textbf{\textit{I}}\textsubscript{truth} from \textbf{\textit{D}} to \textbf{\textit{TV}} (the complementary case 8, for rule implications, is also given, unchanged):
\\[0.6em]
\textbf{\textit{Case 3. Psoa formulas}}: \\[0.1em]
\vspace{-1.2em}
\[
    \tvaltt{f(\sldots)}{} = \ipmap[truth]{\ipmap{f(\sldots)}}
\]
\vspace{-1.6em}
\[
    \tvaltt{o\#f(\sldots)}{} = \ipmap[truth]{\ipmap{o\#f(\sldots)}}
\]
\vspace{-0.5em}

For the
oidful 
formula,
consisting of an object-typing membership, two bags of tuples representing a conjunction of
all the object-centered tuples,
and two bags of slots representing a conjunction of
all the object-centered slots,
the following
\emph{describution} restriction is used, where $\Mathtt{m^{+}},\Mathtt{m^{-}},\Mathtt{k^{+}},\Mathtt{k^{-}}\geq 0$:
\pagebreak
\[
\tvaltt{
\begin{array}{r@{}l}
\Mathtt{o\#f(} & \Mathtt{+[t^{+}_{1,1}\sldots t^{+}_{1,n^{+}_1}]\sldots +\![t^{+}_{m^{+},1}\sldots t^{+}_{m^{+},n^{+}_{m^{+}}}]} \\
& \Mathtt{-[t^{-}_{1,1}\sldots t^{-}_{1,n^{-}_1}]\sldots -\![t^{-}_{m^{-},1}\sldots t^{-}_{m^{-},n^{-}_{m^{-}}}]} \\
& \Mathtt{p^{+}_1\ttarrowdep{}v^{+}_1\sldots p^{+}_{k^{+}}\ttarrowdep{}v^{+}_{k^{+}}} \\
& \Mathtt{p^{-}_1\ttarrow{}v^{-}_1\sldots p^{-}_{k^{-}}\ttarrow{}v^{-}_{k^{-}})}
\end{array}
}{\;t}
\]
if and only if
\[
\begin{array}{clcl}
  & \tvaltt{o\#f}{} \\
= & \tvaltt{o\#f(+[t^{+}_{1,1}\sldots t^{+}_{1,n^{+}_1}])}{} & = \sldots = & \tvaltt{o\#f(+\![t^{+}_{m^{+},1}\sldots t^{+}_{m^{+},n^{+}_{m^{+}}}])}{} \\
= & \tvaltt{o\#Top(-[t^{-}_{1,1}\sldots t^{-}_{1,n^{-}_1}])}{} & = \sldots = & \tvaltt{o\#Top(-\![t^{-}_{m^{-},1}\sldots t^{-}_{m^{-},n^{-}_{m^{-}}}])}{} \\
= & \tvaltt{o\#f(p^{+}_1\ttarrowdep{}v^{+}_1)}{} & = \sldots = & \tvaltt{o\#f(p^{+}_{k^{+}}\ttarrowdep{}v^{+}_{k^{+}})}{} \\
= & \tvaltt{o\#Top(p^{-}_1\ttarrow{}v^{-}_1)}{} & = \sldots = & \tvaltt{o\#Top(p^{-}_{k^{-}}\ttarrow{}v^{-}_{k^{-}})}{} \\
= & \true
\end{array}
\]

On the right-hand side of the ``if and only if'' there are
$1\!+\Mathtt{m^{+}}\!+\Mathtt{m^{-}}\!+\Mathtt{k^{+}}\!+\Mathtt{k^{-}}$ \linebreak
subformulas splitting the left-hand side into:
(1) an object membership; \linebreak
(2) $\Mathtt{m^{+}}$ object-centered tupled formulas, each associating the object and the predicate with a tuple;
(3) $\Mathtt{m^{-}}$ object-centered tupled formulas, each associating the object with a tuple using the root predicate {\tt Top};
(4) $\Mathtt{k^{+}}$ object-centered slotted formulas, each associating the object and the predicate with an attribute-value pair;
and (5) $\Mathtt{k^{-}}$ object-centered slotted formulas, each associating the object with an attribute-value pair using the root predicate {\tt Top}.

To ensure that all members of a subpredicate are also members of its
superpredicates, i.e. \texttt{o\,\#\,f} and
\texttt{f\,\#\#\,g} imply \texttt{o\,\#\,g},
the following \emph{subpredicate-membership} restriction is imposed:
\vspace{-0.6em}
\begin{itemize}
\item
If
\textit{TVal}\calsubscript{I}(\texttt{o\,\#\,f})\,=\,\textit{TVal}\calsubscript{I}(\texttt{f\,\#\#\,g})\,=\,
\textbf{t}  then 
\textit{TVal}\calsubscript{I}(\texttt{o\,\#\,g})\,=\;\textbf{t}.
\end{itemize}
\vspace{0.2em}
\textbf{\textit{Case 8. Rule implication}}: \\[-1.8em]
\begin{itemize}
\item
\textit{TVal}\calsubscript{I}(\textit{conclusion}\;:-\;\textit{condition})\;\;=\;\;\textbf{t} ~ if ~
\textit{TVal}\calsubscript{I}(\textit{conclusion})\;\;=\;\;\textbf{t} \linebreak
\hspace*{21.0em} or
\textit{TVal}\calsubscript{I}(\textit{condition})\;\;=\;\;\textbf{f}.
\item
\textit{TVal}\calsubscript{I}(\textit{conclusion}\;:-\;\textit{condition})\;\;=\;\;\textbf{f} ~ otherwise.
\end{itemize}

\section{PSOA RuleML Translation by PSOATransRun}\label{sec:implement}

To achieve a reference implementation for deduction in PSOA RuleML, we have realized 
the PSOATransRun prototype as an open-source framework system, \linebreak
generally referred to as 
PSOATransRun[{\it translator},{\it runtime}], with a pair of \linebreak
components `plugged in' as parameters to create instantiations \cite{Boley11d,ZouPeter-Paul+12,ZouBoley15}\footnote{\url{http://wiki.ruleml.org/index.php/PSOA_RuleML\#Implementation}}. \linebreak
The {\it translator} component maps KBs and queries from PSOA RuleML to an intermediate language.
The {\it runtime} component executes queries against a KB,
both
in the intermediate language, and extracts the results.
Our focus
is
on translators, reusing the targeted runtime systems as `black boxes'.
Each translator is composed of a chain of transformers, which implement internal translation steps within PSOA RuleML, as well as a converter, which implements an external translation step to the intermediate language.
For our current two instantiations, we have chosen two intermediate languages: the first-order subset, TPTP-FOF, of TPTP \cite{Sutcliffe09}\footnote{TPTP-FOF is also targeted by \url{http://wiki.ruleml.org/index.php/TPTP_RuleML}.} and the Horn-logic subset of ISO Prolog \cite{ISOProlog95}.
Since these are also standard languages, their translator components in PSOATransRun serve both for PSOA RuleML implementation and interoperation \cite{ZouBoley15}. 

The chain targeting TPTP requires four PSOA-internal translation steps \linebreak
-- unnesting,
subclass\footnote{In PSOATransRun software/papers, ``subclass'' is pars pro toto for ``subpredicate''.} rewriting,
objectification, and describution -- while
the chain into ISO Prolog requires three subsequent translation steps -- Skolemization,
conjunctive-conclusion splitting,
and flattening -- since ISO Prolog has lower expressivity
(e.g., requiring head existentials to be eliminated via Skolemization).

To
realize
the perspectival knowledge of the
PSOA RuleML language since Version 1.0
in
the PSOATransRun system since Version 1.3, the transformation step of describution is revised to
replace
every oidful psoa atom having the general form
\[
\begin{array}{r@{}l}
\Mathtt{o\#f(} & \Mathtt{+[t^{+}_{1,1}\sldots t^{+}_{1,n^{+}_1}]\sldots +\![t^{+}_{m^{+},1}\sldots t^{+}_{m^{+},n^{+}_{m^{+}}}]} \\
& \Mathtt{-[t^{-}_{1,1}\sldots t^{-}_{1,n^{-}_1}]\sldots -\![t^{-}_{m^{-},1}\sldots t^{-}_{m^{-},n^{-}_{m^{-}}}]} \\
& \Mathtt{p^{+}_1\ttarrowdep{}v^{+}_1\sldots p^{+}_{k^{+}}\ttarrowdep{}v^{+}_{k^{+}}} \\
& \Mathtt{p^{-}_1\ttarrow{}v^{-}_1\sldots p^{-}_{k^{-}}\ttarrow{}v^{-}_{k^{-}})}
\end{array}
\]
-- reflecting the describution restriction of Section~\ref{sec:semantics} -- with the conjunction
\begin{align*}
\begin{split}
\Mathtt{And(} & \Mathtt{o\#f} \\
& \Mathtt{o\#f(+[t^{+}_{1,1}\sldots t^{+}_{1,n^{+}_1}]) \sldots o\#f(+[t^{+}_{m^{+},1}\sldots t^{+}_{m^{+},n^{+}_{m^{+}}}])} \\
& \Mathtt{o\#Top(-\![t^{-}_{m^{-},1}\sldots t^{-}_{m^{-},n^{-}_{m^{-}}}]) \sldots o\#Top(-[t^{-}_{m^{-},1}\sldots t^{-}_{m^{-},n^{-}_{m^{-}}}])} \\
& \Mathtt{o\#f(p^{+}_1\ttarrowdep{}v^{+}_1) \sldots o\#f(p^{+}_{k^{+}}\ttarrowdep{}v^{+}_{k^{+}})} \\
& \Mathtt{o\#Top(p^{-}_1\ttarrow{}v^{-}_1) \sldots o\#Top(p^{-}_{k^{-}}\ttarrow{}v^{-}_{k^{-}}))}
\end{split}
\end{align*}

Examples of the transformation have already been given in Section~\ref{unscopingtodescribution}.

The describution-yielded conjuncts are converted to Prolog and TPTP, which share the same syntax for atoms.
This conversion uses the reserved runtime predicates {\tt memterm}, {\tt tupterm}, {\tt prdtupterm}, {\tt sloterm}, and {\tt prdsloterm} for, respectively, membership, independent-tuple, dependent-tuple, 
independent-slot, and dependent-slot terms, as shown in the following 
table, where \mapprl{} denotes the recursive mapping from PSOA to Prolog or TPTP. The predicates {\tt memterm}, {\tt tupterm}, and {\tt sloterm} have been used since our previous work~\cite{ZouBoley15} while {\tt prdtupterm} and {\tt prdsloterm} are newly introduced to translate dependent descriptors.

\begin{center}
{\tt
\begin{tabular}{|@{\ }c@{\ }|@{\ }c@{\ }|}
    \hline
    {\bf\rm Psoa Atoms} & {\bf\rm Prolog and TPTP Atoms} \\ \hline
    o\#f & memterm(\mapprl{\tt o},\mapprl{\tt f}) \\ \hline
    o\#Top(-[\tup]) & tupterm(\mapprl{\tt o},\mapprl{\tt t_1},\sldots,\mapprl{\tt t_n}) \\ \hline
    o\#f(+[\tup]) & prdtupterm(\mapprl{\tt o},\mapprl{\tt f},\mapprl{\tt t_1},\sldots,\mapprl{\tt t_n}) \\ \hline
    o\#Top(p->v) & sloterm(\mapprl{\tt o},\,\mapprl{\tt p},\,\mapprl{\tt v}) \\ \hline
    o\#f(p+>v) & prdsloterm(\mapprl{\tt o},\,\mapprl{\tt f},\,\mapprl{\tt p},\,\mapprl{\tt v}) \\ \hline
\end{tabular}
}
\end{center}

All these `machine' predicates are oidless while taking the \mapprl{}-mapped OID {\tt o} as
their
first
argument.
Also, {\tt memterm}, {\tt prdtupterm}, and {\tt prdsloterm} take the mapped predicate {\tt f} as the second argument.
Moreover, {\tt tupterm} and {\tt prdtupterm} take the {\tt n} mapped components of the tuple as the remaining arguments.
Finally, {\tt sloterm} and {\tt prdsloterm} take the mapped slot name and the mapped slot filler as the last two arguments.
The dependence-encoding {\tt prdtupterm} and {\tt prdsloterm}
are
extensions of, respectively, the independence-encoding {\tt tupterm} and {\tt sloterm} with an extra predicate argument \mapprl{\tt f}.
The extension of the three earlier runtime predicates by the two new ones
does not incur
an overhead when not used and -- as demonstrated below -- can speed up execution when used. 

In~\cite{ZouBoley16}, we introduced static/dynamic objectification as an alternative to static objectification in~\cite{Boley11d}. 
The static/dynamic objectification tries to avoid generating explicit static OIDs for Prolog-like relations, instead constructing dynamic virtual OIDs at query time if and when bindings for OID variables are
requested.
The dynamic part of static/dynamic objectification, i.e. {\em dynamic objectification}, applies to atoms having a \textcolor{orange}{{\em relational predicate}} in a given KB, 
which
\textcolor{orange}{was defined} as a predicate that has no occurrence in an oidful, multi-tuple, or slotted atom.
Equivalently, a relational predicate
was to occur
only in oidless atoms that are empty or have one tuple.
With
the new
dependent/independent atoms
and empty atoms now differing from atoms having one dependent empty tuple,
a \linebreak
\textcolor{red}{{\em relational predicate}}
\textcolor{red}{is further restricted} to
a predicate
with no
occurrence in an \linebreak
oidful, empty,
independent-tuple-ful,
multi-tuple, or slotted atom.
Equivalently, it occurs only in oidless atoms that have one dependent tuple.
For an atom having an independent tuple
the tuple is intended to become separated from the predicate via the atom's describution.
Since dynamic objectification is designed to keep the predicate together with the tuple, it is not suitable for such an atom.

To explore performance trade-offs for differently modeled KBs,
in a series of experiments we measured the runtime of 
tupled vs. slotted and dependent vs. independent
variations of rule-chaining test cases, Chain, in PSOATransRun 1.3.1's Prolog instantiation, which -- via the InterProlog API\footnote{\url{http://interprolog.com}} -- employs XSB Prolog as the underlying engine.
The experiments were conducted 
with a standard XSB 3.7 installation on Ubuntu 11 running on a 
VirtualBox 4.3.16 virtual machine with 4GB memory over a Windows 7 host on an Intel Core i7-2670QM 2.20GHz CPU.
Since PSOA RuleML's main area of differentiation is in offering novel kinds of atoms, as systematized in the metamodel of Appendix~\ref{sec:psoametamodel}, we focus the discussion on test-querying single (rather than 
conjunctions/joins of) atoms through rule chains of increasing lengths (while
various other test cases -- some with conjunctive queries -- are provided online\footnote{\url{http://wiki.ruleml.org/index.php/PSOA_RuleML\#Test_Cases}}).

We used Python-based generators to create four groups of Chain test cases,\footnote{The programs and tests
are online at \url{http://psoa.ruleml.org/testcases/chain/}.
}
each probing one of the four major kinds of atoms: dependent-tuple, independent-tuple, dependent-slot, and independent-slot.
Each group has test cases distinguished by the number $k$ of KB rules, which is a parameter of the group's \linebreak
generator (detailed below).
Each generated test case includes one KB and one query of the same dependency kind (enabling successful query answering). \linebreak
For each test case, the KB consists of one fact and $k\geq 0$ rules.

In the dependent-tuple group, each generated KB consists of the fact \linebreak
{\tt \_r0(\_a1 \_a2 \_a3)} (an abbreviation of {\tt \_r0(+[\_a1 \_a2 \_a3])}) and $k$ rules
of the following form ($i=1,\ldots,k,i'=i-1$): \\[5pt]
{\tt
\begin{tabular}{@{}l@{}l}
Fo & rall ?X1 ?X2 ?X3 ( \\
    & \_r$i$(?X1 ?X2 ?X3)\,:-\,\_r$i'$(?X1 ?X2 ?X3) \\
) \\[5pt]
\end{tabular}
} \\
\mycomment{
\begin{center}
{\tt 
Forall\ ?X1 ?X2 ?X3 (\ \_r$i$(?X1 ?X2 ?X3)\,:-\,\_r$i'$(?X1 ?X2 ?X3))
}
\end{center}
}
The dependent-tuple query of the form {\tt \_r$k$(?X1 ?X2 ?X3)}, posed to this $k$-rule KB, has one answer, {\tt ?X1=\_a1,?X2=\_a2,?X3=\_a3}.

In the dependent-slot group, each KB consists of one fact {\tt \_r0(\_p1+>\_a1 \_p2+>\_a2 \_p3+>\_a3)} and $k$ rules
of the following form ($i=1,\ldots,k,i'=i-1$): \\[5pt]
{\tt
\begin{tabular}{@{}l@{}l}
Fo & rall ?X1 ?X2 ?X3 ( \\
   & \_r$i$(\_p1+>?X1 \_p2+>?X2\ \_p3+>?X3)\,:-\,\_r$i'$(\_p1+>?X1 \_p2+>?X2 \_p3+>?X3) \\
) \\[5pt]
\end{tabular}
}
\mycomment{
\begin{center}
{\tt 
Forall\ ?X1 ?X2 ?X3 (\ \_r$i$(\_p1+>?X1 \_p2+>?X2\ \_p3+>?X3)\,:-\,\_r$i'$(\_p1+>?X1 \_p2+>?X2\ \_p3+>?X3))
}
\end{center}
}
The dependent-slot query {\tt \_r$k$(\_p1+>?X1 \_p2+>?X2 \_p3+>?X3)}, posed to this $k$-rule KB, has the same answer, {\tt ?X1=\_a1,?X2=\_a2,?X3=\_a3}.

The dependent-slot group can be seen as a dependency-preserving, positional-to-slotted-reduced version of the dependent-tuple group (cf. Section~\ref{dependencetranslations}).

The independent-tuple and independent-slot groups are constructed by toggling the two dependent groups (cf. Footnote~\ref{footnote:deptoindep}).

\mycomment{
The KB has four variants, each consisting of atoms of one kind, having (1) a dependent length-three tuple ({\tt \_r$i$(+[?X1 ?X2 ?X3])}, abbreviated to {\tt \_r$i$(?X1 ?X2 ?X3)}), (2) an independent length-three tuple ({\tt \_r$i$(-[?X1 ?X2 ?X3])}),
(3) three dependent slots ({\tt \_r$i$(\_p1+>?X1 \_p2+>?X2 \_p3+>?X3)}), or (4) three independent slots ({\tt \_r$i$(\_p1->?X1 \_p2->?X2 \_p3->?X3)}).
}

Starting with $k$=0 rules and increasing in steps of 50 rules until reaching $k$=500 rules,
we generated eleven test cases for each group and measured their query execution time.
For the dependent-tuple group, we also compared the query execution time using a switch in PSOATransRun between the above-discussed static vs. static/dynamic objectification.
For the other three groups, where none of the predicates can be relational, static/dynamic objectification degenerates to static objectification, hence we
did not compare the two settings.

\begin{figure}[htp]
\centering
\includegraphics[width=.5\textwidth]{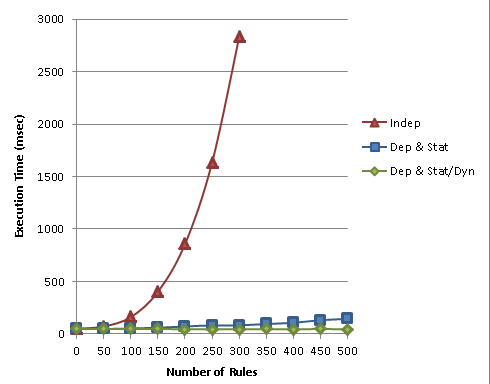}
\caption{Execution time of eleven Tupled Chain test cases.}
\label{fig:tupled-chain}
\end{figure}

\begin{figure}[htp]
\centering
\includegraphics[width=.5\textwidth]{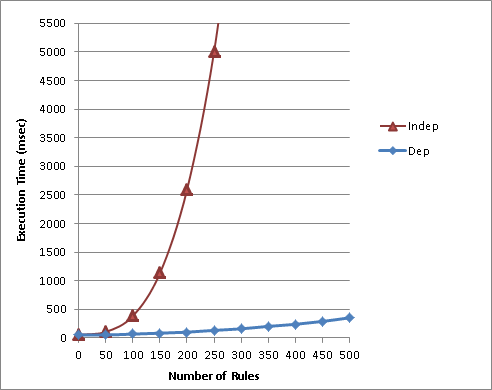}
\caption{Execution time of eleven Slotted Chain test cases.}
\label{fig:slotted-chain}
\end{figure}

The results for the tupled groups are shown in Table~\ref{tbl:tupled-chain} and Fig.~\ref{fig:tupled-chain} while the results for the slotted groups are shown in Table~\ref{tbl:slotted-chain} and Fig.~\ref{fig:slotted-chain}.
In the tables, the shortcut ``\textsl{query-err}'' means that the query execution
ran out of memory in XSB Prolog.

\newcolumntype{Y}{>{\Centering}X}
\begin{table}[htp]
    \centering
    \caption{Execution time of eleven Tupled Chain test cases.}
    \label{tbl:tupled-chain}
    \begin{tabularx}{.65\textwidth}{c|c|Y|Y|Y|}
         \multicolumn{2}{c|}{} & \multicolumn{3}{c|}{Dependency \& Objectification Choices}  \\ \cline{3-5}
         \multicolumn{2}{c|}{} & Indep & \multicolumn{2}{c|}{Dep} \\ \cline{4-5}
         \multicolumn{2}{c|}{} & & Stat & Stat/Dyn \\ \hline 
         & 0 & 47 & 51 & 49 \\ \cline{2-5}
         & 50 & 72 & 47 & 47 \\ \cline{2-5}
         & 100 & 161 & 52 & 46 \\ \cline{2-5}
         & 150 & 403 & 59 & 48 \\ \cline{2-5}
        Number of & 200 & 858 & 71 & 44 \\ \cline{2-5}
        Rules & 250 & 1636 & 81 & 44 \\ \cline{2-5}
         & 300 & 2834 & 82 & 45 \\ \cline{2-5}
         & 350 &  & 95 & 46 \\ \cline{2-2}\cline{4-5}
         & 400 & \textsl{query-err} & 106 & 44 \\ \cline{2-2}\cline{4-5}
         & 450 &  & 131 & 47 \\ \cline{2-2}\cline{4-5}
         & 500 &  & 143 & 44 \\ \hline
    \end{tabularx}
\end{table}
\begin{table}[htp]
    \centering
    \caption{Execution time of eleven Slotted Chain test cases.}
    \label{tbl:slotted-chain}
    \begin{tabularx}{.5\textwidth}{c|c|Y|Y|}
         \multicolumn{2}{c|}{} & \multicolumn{2}{c|}{Dependency Choices}  \\ \cline{3-4}
         \multicolumn{2}{c|}{} & Indep & Dep \\ \hline
         & 0 & 55 & 54 \\ \cline{2-4} 
         & 50 & 106 & 52 \\ \cline{2-4} 
         & 100 & 384 & 67 \\ \cline{2-4} 
         & 150 & 1134 & 83 \\ \cline{2-4} 
        Number of & 200 & 2595 & 101 \\ \cline{2-4} 
        Rules & 250 & 5012 & 132 \\ \cline{2-4} 
         & 300 & 8613 & 160 \\ \cline{2-4} 
         & 350 &  & 202 \\ \cline{2-2}\cline{4-4} 
         & 400 & \textsl{query-err} & 239 \\ \cline{2-2}\cline{4-4} 
         & 450 &  & 289 \\ \cline{2-2}\cline{4-4} 
         & 500 &  & 352 \\ \hline
    \end{tabularx}
\end{table}

From the tables and figures, we can see that the slotted test cases are slower than their tupled counterparts.
This is
because each slotted atom has three slots while each tupled atom has one tuple:
hence, after describution, each slotted atom becomes a 4-ary conjunction while each tupled atom becomes a 2-ary conjunction,
leading to more branches for the slotted versions during reasoning.

Also from the tables and figures, the test cases using independent descriptors are slower than their dependent counterparts.
This is because the $k$ rules in the Chain test cases
differ only in their predicates,
yet
for independent descriptors,
describution separates the predicate from the descriptors, leaving behind {\tt Top}-typed, single-descriptor atoms in rule conclusions and conditions that
can be unified with each other, leading to a significant increase in reasoning time.

For the above and similar dependent-tuple tests,
static/dynamic objectification is faster than static objectification since the former keeps the PSOA relationships in Chain,
converting
them directly
to Prolog relationships,
while the latter introduces explicit (Skolem-function-nesting) OIDs for the relationships,
descributes them into conjunctions,
and translates them via reserved predicates.

These experiments
indicate
that: (1)
for rules whose conclusions and conditions
contain atoms with different predicates but unifiable descriptors,
dependent modeling of those descriptors is more efficient
than their independent modeling;
(2) for argument collections
that occur jointly in many atoms
(e.g., arguments {\tt ?X1}, {\tt ?X2}, {\tt ?X3} in Chain), tupled modeling is more efficient than slotted modeling.


\vspace{-0.1in}
\section{Conclusions}\label{sec:conc}

PSOA RuleML, since Version 1.0 featuring perspectival knowledge, constitutes a succinct yet expressive language mostly due to its orthogonal overall design according to dimensions D$_1$-D$_3$ of our metamodel for atoms in Fig.~\ref{Metamodel}.
Dependency, the novel independent/dependent dimension D$_3$, is 
defined via independent/ dependent descriptors,
which can be tuples or slots. 
Perspectival knowledge is \linebreak
illustrated by the Rich TA example, visualized in  Grailog in Fig.~\ref{TAexample}, describing the same OID individual differently under different perspectives using atoms having different predicates.
Each descriptor of an atom can be independent or dependent from the predicate, which, respectively, allows or disallows the rescoping of the 
descriptor to a different predicate for a given OID.
Enabling descriptor inheritance, default descriptors are realized by default rules and facts.

To
enable
perspectival knowledge, the presentation and serialization syntaxes of PSOA RuleML are appropriately augmented and the model-theoretic semantics is revised in Version 1.0.
The novel
D$_3$
was first realized in
PSOATransRun 1.3, 
whose translator component revises (a) the multiple PSOA-internal translation steps, focused on the tupribution/slotribution -- i.e., describution -- step and 
(b) the conversion to Prolog or TPTP. The conversion uses new reserved runtime predicates {\tt prdtupterm} and {\tt prdsloterm} for dependent descriptors.

\setlength{\baselineskip}{1.15em}
The complete earlier Grailog visualization systematics for psoa atoms of~\cite{Boley15b}, \linebreak
introduced before the independent/dependent distinction, has been extended to the dependency dimension according to Fig.~\ref{TAexample}, where
an oidless atom can \linebreak
-- instead of a rectangular OID box -- use a `box' degenerated to a ``branch line''~\cite{Boley15b}, \linebreak
e.g. as a starting point for descriptor (hyper)arcs~\cite{Boley19c}.

\setlength{\baselineskip}{1.15em}
Future work on PSOA RuleML is partly driven by the structured agenda for PSOATransRun\footnote{\url{http://wiki.ruleml.org/index.php/PSOATransRun_Development_Agenda}}, with open-source development organized via GitHub.

This may involve the metamodel's dimension $\mathrm{D_0}$.
Its two
bags of descriptors
could be combined to one bag with cardinality $\mathrm{m+k}$.
Conversely, as in the semantics, $\mathrm{D_0}$'s two bags could be refined to four bags,
also distinguishing independent vs. dependent atoms,
enabling another mapping:
$\mathrm{D_0}\rightarrow\mathrm{D_3}$.
The current 2-bag $\mathrm{D_0}$ can be reconstructed from the 4-bag $\mathrm{D_0}$ by $\mathrm{m^{+} +m^{-}}$ and $\mathrm{k^{+} + k^{-}}$
(also, the 1-bag $\mathrm{D_0}$ directly from the 4-bag $\mathrm{D_0}$ by $\mathrm{m^{+} +m^{-} + k^{+} + k^{-}}$).

\setlength{\baselineskip}{1.15em}
The schema specification of the PSOA RuleML/XML 1.03 serialization syntax in Relax NG\footnote{\url{http://wiki.ruleml.org/index.php/PSOA_RuleML\#Syntax}} (allowing automatic translation to XSD) should be refined. 
The preliminary RuleML/JSON syntax\footnote{\url{http://wiki.ruleml.org/index.php/RuleML_in_JSON}}
can be adapted to PSOA RuleML 1.03.

\setlength{\baselineskip}{1.15em}
Use cases employing PSOA RuleML have been conducted, e.g. for data querying and mapping in the geospatial \cite{Zou15} and biomedical domains \cite{AlManirRiazanov+16}.
Further PSOA RuleML 1.03 KBs should be developed, 
e.g. with (legal/regulatory) knowledge about ships (cf. the Port Clearance Rules use case~\cite{ZouBoley+17}) and cars, where, e.g., amphibious vehicles are calling for perspectival knowledge.
Such use cases are being collected on the PSOA RuleML wiki page\footnote{\url{http://wiki.ruleml.org/index.php/PSOA_RuleML\#Use_Cases}}.

\setlength{\baselineskip}{1.15em}
Expanding on Section~\ref{sec:inheritance},
the well-known ``Nixon Diamond'' problem~\cite{Horty2012} can also be modeled as perspectival knowledge in a PSOA KB:
(1) {\tt Quaker\{policy+> 
pacifist\}}, i.e. under the perspective of being a {\tt Quaker}, one's {\tt policy} is {\tt pacifist}.
(2) {\tt Republican\{policy+>nonpacifist\}}, i.e. as a {\tt Republican}, one's {\tt policy} is {\tt nonpacifist}.
(3) Nixon is both a {\tt Quaker} and a {\tt Republican}, leading to a conflict.
In this modeling, querying the {\tt policy} slot (fillers: {\tt pacifist} or {\tt nonpacifist}) without specifying the perspective (predicates: {\tt Quaker} or {\tt Republican}) would fail.
Thus, the semantics
is similar to the ``skeptical'' approach, where no conflicted conclusions can be drawn. 
In contrast, using non-perspectival modeling ({\tt Quaker\{policy->pacifist\}} and {\tt Republican\{policy->nonpacifist\}}), the \linebreak
{\tt policy} slot would be predicate-independent, and the semantics
similar to the ``credulous'' approach, where querying the {\tt policy} slot would give conflicted ({\tt pacifist} and {\tt nonpacifist}) conclusions.

\mycomment{
\begin{verbatim}
Forall ?o (
  ?o#Republican(policy+>nonpacifist) :- ?o#Republican
)
Forall ?o (
  ?o#Quaker(policy+>pacifist) :- ?o#Quaker
)
Nixon#Republican
Nixon#Quaker
\end{verbatim}

The rules could be shortcut to sth. like
Republican{policy+>nonpacifist}, statically expanding to the those rules.

\begin{verbatim}
Forall ?o (
  ?o#Top(policy->nonpacifist) :- ?o#Republican
)
Forall ?o (
  ?o#Top(policy->pacifist) :- ?o#Quaker
)
Nixon#Republican
Nixon#Quaker
\end{verbatim}
}

\setlength{\baselineskip}{1.15em}
Moreover, the connections between
contextual and perspectival knowledge can be further elaborated, including (mutual) reductions:
Besides the reductions discussed in
Section~\ref{dependencetranslations},
perspectival knowledge could be emulated via contextual knowledge
by contextualizing the clauses describing the same global (IRI) OID\footnote{Copies of the same local (``{\tt \_}''-prefixed) OID in different (perspectival) contexts would be renamed apart on merging, which would usually be unintended.} -- in a very fine-grained manner -- w.r.t.
their different predicates, permitting dependent descriptors to become independent. In particular, for the OID {\tt John} a context for each of the three predicates {\tt Teacher}, {\tt TA}, and {\tt Student} could be created (an OID's multi-membership becomes a `multi-contextship'), where, e.g., the {\tt Teacher} context would permit independent descriptors like \texttt{dept->Physics}.
Conversely, for a context-partitioned KB of oidless ground facts, OIDs could be introduced to represent context names
(similar to, e.g., ``named graphs''~\cite{RDFConcepts14}), 
where the OID-typing predicates could provide cross-contextual perspectives.

\setlength{\baselineskip}{1.15em}
The PSOA RuleML
reference implementation PSOATransRun 1.4.2 
should be further developed as part of the PSOATransRun framework, whose instantiations target both Prolog (with Naf) and TPTP.
The performance of the Prolog instantiation -- e.g.,
based on
Tables~\ref{tbl:tupled-chain} and \ref{tbl:slotted-chain},
and on PSOA user feedback -- 
should be further increased as part of the next release. 
Besides accepting the presentation syntax,
it should also accept the serialization syntax,
and permit translations between the two
(for the serialization-to-presentation direction using the PSOA RuleML API\footnote{\url{http://wiki.ruleml.org/index.php/PSOA_RuleML_API}}). 
Since, as shown in this paper, much of the expressivity of perspectival knowledge representation can already
be realized on the function-free level of Datalog (rather than requiring Horn logic), a
new
instantiation of PSOATransRun should be done for Datalog PSOA (specializing the current Hornlog PSOA) by targeting an (object-relational) database engine,
whose ``views''  implement rules.
Conversely, PSOA's atoms could be \texttt{<repo>}-tuple-extended~\cite{Boley15b} and carried up to Hornlog+, FOL, and all other levels of Deliberation RuleML and, in PSOA Prova~\cite{GraetzBoley+18}, to Reaction RuleML etc. 

\setlength{\baselineskip}{1.15em}
Finally, some or all of the dimensions of the PSOA metamodel could be transferred to other object-centered logics and deductive database systems.

\vspace{-0.1in}
\section{Acknowledgements}
\setlength{\baselineskip}{1.15em}
{\small
Initial feedback, on PSOA RuleML 1.0, was obtained in RuleML+RR 2017's Standards Session\footnote{\url{http://2017.ruleml-rr.org/schedule/}}.
We thank Jon Hael Simon Brenas, Alexandre Riazanov, Tara Athan, Sadnan Al Manir, and Theodoros Mitsikas for helpful comments on this paper.
Thanks go to: Sofia Almpani, for a use case on medical devices; Theodoros Mitsikas, for a use case on air traffic control and for PSOATransRun releases since 1.3.2-b, including an SWI Prolog adaptation; Miguel Calejo, for an SWI InterProlog upgrade.
Also to Tara Athan and Rima Chaudhari, for enabling a PSOA RuleML/XML 1.03 release.
NSERC is thanked for 
Discovery Grants.
}

\vspace{-0.1in}


\bibliographystyle{splncs}

\bibliography{PSOAPerspectivalKnowledge}

\begin{appendix}
\section{The Metamodel for Psoa Atoms of PSOA RuleML}
\label{sec:psoametamodel}

{\small
This appendix introduces the metamodel for PSOA RuleML 1.03's psoa atoms. \linebreak
Psoa atoms can be initially characterized using a \textbf{quantitative} dimension D$_0$.
This zeroth dimension classifies an atom via its zero-or-more ($\geq$0) descriptors partitioned into bags of m ($\geq$0) tuples and k ($\geq$0) slots, where both bags can have, 
e.g., zero (=0), zero-or-more ($\geq$0), single (=1), one-or-more ($\geq$1), or multiple ($\geq$2) descriptors (for details and the concrete syntax see Section~\ref{sec:presentationsyntax}). 
E.g., empty atoms are characterized by D$_0$(m=0,k=0), i.e. have neither tuples (m=0) nor slots (k=0),
but like all atoms must have a predicate (which can be the {\it root predicate} {\tt Top}) and may have an OID.
Empty atoms constitute a category of their own since 
being {\it descriptorless} (tupleless and slotless) will make the 
tupled/slotted\footnote{To emphasize the option of multiple tuples, the earlier terms ``positional/slotted'' have been replaced by ``tupled/slotted'' in most cases.} and independent/dependent 
distinctions
non-applicable
to them (not even expressible in the syntax of Section~\ref{sec:presentationsyntax}).
For non-empty atoms, D$_0$ is used to define D$_2$, as shown by the D$_0$ $\rightarrow$ D$_2$ mapping in Fig.~\ref{Metamodel}.

There are three orthogonal \textbf{qualitative} dimensions D$_1$-D$_3$ for non-empty atoms generating eighteen subcubes.
Refining the earlier six quadrants of the ``psoa table''~\cite{Boley15b}, the PSOA 1.03 subcubes are labeled according to three layers of six, for non-empty atoms that are
\textbf{in}dependent, \textbf{de}pendent, and \textbf{i}ndependent+\textbf{d}ependent,
as well as suffixed with digits 1\,--\,6 in each layer.
Besides all having systematic labels/digits, the layer-in{} and -de{} subcubes have common names such as
in{}4, i.e. oidful, one-or-more-slotted, independent atoms,
referred to as {\it framepoints};
other subcubes are further specialized by D$_0$
for defining subsets, such as de{}1, which is D$_0$(m=1,k=0)-specialized to oidless, single-tupled, dependent atoms, having the common name {\it relationships}.\footnote{The subcubes partition the set of PSOA RuleML languages into languages of, e.g., relationship and framepoint facts and/or rules, some of which could have anchor names, as introduced for RuleML/XML~\cite{AthanBoley14}.}

\setlength{\baselineskip}{1.13em}
In each layer, atoms are characterized using the same two distinctions.
The first \linebreak
dimension D$_1$ distinguishes atoms that are oidless-vs.-oidful predicate applications.
The second dimension D$_2$ distinguishes atoms having as descriptors one or more tuples vs. one or more slots vs. combining one or more tuples plus one or more slots.
The two main quadrants of the earlier psoa table are also accommodated by these dimensions via
the above-mentioned de{}1 subcube (specializing to relationships) and in{}4 subcube (constituting framepoints).
Intuitively speaking, because a tuple contains zero or more elements, a relationship affords only a single (m=1) descriptor; because a plain slot pairs a name with only a plain filler,\footnote{Here, ``plain'' refers to an ordinary slot with a non-tuple-valued filler; cf. Footnote~\ref{footnote:prop}.} a framepoint affords one or more (k$\geq$1) descriptors. \linebreak
Similarly, because of the only tuple's (non-association with an OID but) dependence on a predicate, relationships are dependent; because of the one or more slots' (association with an OID but) independence from a predicate, framepoints are independent.

In this paper, dimensions D$_0$-D$_2$ are augmented by the dimension D$_3$ of
atoms being {\it \textbf{in}dependent}, i.e. \textcolor{violet}{only having (one or more)} predicate-independent descriptors, vs.
{\it \textbf{de}pendent}, i.e. \textcolor{violet}{only having (one or more)} predicate-dependent descriptors, vs.
{\it \textbf{i}ndependent+\textbf{d}ependent}, i.e. 
\textcolor{violet}{combining (one or more)} predicate-independent \textcolor{violet}{plus (one or more)} predicate-dependent descriptors.\footnote{The dependency dimension D$_3$  for an atom is thus based on the dependency distinction (independent/dependent or independence/dependence) for its descriptors.
With the dependency superscripts
(``$^{-}$''/``$^{\Mathtt{+}}$'')
of Section \ref{sec:presentationsyntax}: 
independent iff $\mathrm{m^{-} + k^{-}}$ $\geq$ 1 and $\mathrm{m^{+} + k^{+}}$ = 0;
dependent iff $\mathrm{m^{+} + k^{+}}$ $\geq$ 1 and $\mathrm{m^{-} + k^{-}}$ = 0;
independent+dependent iff $\mathrm{m^{-} + k^{-}}$ $\geq$ 1 and $\mathrm{m^{+} + k^{+}}$ $\geq$ 1.}
In the systematics of Fig.~\ref{Metamodel}, the zeroth dimension is indicated by (m,k) in/equality pairs,
the first and second dimensions are constituted by the columns and rows of each layer,
and the third dimension is unraveled layer-wise.
As provided above, relationships belong to the dependent layer,
while framepoints belong to the independent layer.
Conversely, \textit{shelfships} are relationship-like oidless, tupled, independent atoms (in{}1), and \textit{pairpoints} are framepoint-like oidful, slotted, dependent atoms (de{}4). 
Also, oidful, tupled+slotted, independent atoms (in{}6) D$_0$(m=1,k$\geq$1)-specialize 
to \textit{shelframepoints};
atoms in (de{}5) D$_0$(m=1,k$\geq$1)-specialize to \textit{relpairships}. 
Analogously to the third rows for the tupled+slotted combination in D$_2$, the third layer 
is introduced for the independent+dependent combination in D$_3$,
allowing
atoms having at least one independent and at least one dependent descriptor.


Collections of atoms broader than one subcube (which represents a basic category) \linebreak
can be specified by just omitting constraints for some dimensions.
For example, \linebreak
omitting all constraints but one, {\it single-tuple atoms} specify all
oidless or oidful, \linebreak
m=1 tuple, k$\geq$0 slot, independent or dependent atoms.
Further non-basic \linebreak
categories of atoms can be constructed as the union of basic or non-basic categories. \linebreak
In particular,
{\it oidless, m=1
tuple, k$\geq$0 slot, dependent atoms}
can be constructed as the union of relationship and relpairship atoms.

The dimensions of the metamodel allow the following categorization of the three color-grouped psoa atoms in Section~\ref{intro}, Fig.~\ref{TAexample}, all of which fixing D$_1$: oidful. 
\vspace{-0.2em}
\begin{itemize}
\item The (red) {\tt Teacher} atom
fixes \ D$_0$(m=1,k=3);
\ D$_2$: tupled+slotted; \newline
D$_3$:
independent+dependent
\item The (green) {\tt Student} atom
fixes \ D$_0$(m=2,k=2);
\ D$_2$: tupled+slotted; \newline
D$_3$: independent+dependent
\item The (brown) {\tt TA} atom
fixes \ D$_0$(m=0,k=1), a single-descriptor case, i.e. \newline
Section~\ref{intro}'s methods (i) and (ii) coincide;
\ D$_2$: slotted;
\ D$_3$: dependent 
\end{itemize}
\vspace{-0.075em}
Such categorizations exemplify PSOA's
novel distinction of (predicate-)independent vs. 
(predicate-)dependent descriptors (tuples and slots) as dimension D$_3$ within the larger design space generated by dimensions D$_0$-D$_3$.

\vspace{0.04cm}
\begin{figure*}
\mycomment{
O L D:
\begin{small}
\begin{center}
  \begin{tabular}{ l | l | l |}
    \textbf{D$_3$: independent} & \ D$_1$: oidless \ & \ D$_1$: oidful \ \\ \hline
    D$_2$: tupled & \ in{}1. & \ in{}2. \\ \hline
    D$_2$: slotted & \ in{}3.  D$_0$(m=0,k$\geq$0): pairships & \ in{}4. D$_0$(m=0,k$\geq$0): \textbf{\textit{framepoints}} \\ \hline
    D$_2$: tupled+slotted \ & \ in{}5.  & \ in{}6. D$_0$(m=1,k$\geq$1): shelframepoints \\ \hline
  \end{tabular}
\end{center}

\begin{center}
  \begin{tabular}{ l | l | l |}
    \textbf{D$_3$: dependent} & \ D$_1$: oidless \ & \ D$_1$: oidful \ \\ \hline
    D$_2$: tupled & \ de{}1. D$_0$(m=1,k=0): \textbf{\textit{relationships}} & \ de{}2. D$_0$(m=1,k=0): shelfpoints \\ \hline
    D$_2$: slotted & \ de{}3. & \ de{}4. \\ \hline
    D$_2$: tupled+slotted \ & \ de{}5. D$_0$(m=1,k$\geq$1): relpairships & \ de{}6.  \\ \hline
  \end{tabular}
\end{center}

\begin{center}
  \begin{tabular}{ l | l | l |}
    \textbf{D$_3$: independent+dependent} & \ D$_1$: oidless \ & \ D$_1$: oidful \ \\ \hline
    D$_2$: tupled & \ id{}1. & \ id{}2. \\ \hline
    D$_2$: slotted & \ id{}3. & \ id{}4. \\ \hline
    D$_2$: tupled+slotted \ & \ id{}5. & \ id{}6. \\ \hline
  \end{tabular}
\end{center}
\end{small}

N E W:
}
\begin{small}
\begin{tabular}{@{}lll}
\textbf{Origin:} \\
D$_0$(m=0,k=0) & \} & descriptorless \\[5pt]

\textbf{Mapping:} \\
\begin{tabular}{@{}l}
D$_0$(m$\geq$1,k=0) $\longrightarrow$ D$_2$: tupled    \\
D$_0$(m=0,k$\geq$1) $\longrightarrow$ D$_2$: slotted \\
D$_0$(m$\geq$1,k$\geq$1) $\longrightarrow$ D$_2$: tupled+slotted
\end{tabular} &
\Bigg\} & descriptorful \\
\end{tabular}

\vspace{0.05em}
\mycomment{
D$_0$(m$\geq$1,k=0) $\longrightarrow$ D$_2$: tupled ---------------------

D$_0$(m=0,k$\geq$1) $\longrightarrow$ D$_2$: slotted ----------------------------- descriptorful

D$_0$(m$\geq$1,k$\geq$1) $\longrightarrow$ D$_2$: tupled+slotted -----------

\begin{array}{l}
\mathrm{D_0(m\geq 1,k=0) \longrightarrow D_1: tupled} \\
\mathrm{D_0(m=0,k\geq 1) \longrightarrow D_1: slotted} \\
\mathrm{D_0(m\geq 1,k\geq 1) \longrightarrow D_1: tupled+slotted}
\end{array}
}

\vspace{0.2cm}
Empty atoms are descriptorless, either oidless or (for memberships) oidful.

Non-empty atoms are constituted as descriptorful by the three layers below.
\\[0.4cm]
  \begin{tabular}{ l | l | l |}
    \textbf{D$_3$: independent\ } & \ D$_1$: oidless \ & \ D$_1$: oidful \ \\ \hline
    D$_2$: tupled & \ in{}1. D$_0$(m=1,k=0): shelfships & \ in{}2. D$_0$(m=1,k=0): shelfpoints \\ \hline
    D$_2$: slotted & \ in{}3: frameships & \ in{}4: \textbf{\textit{framepoints}} \\ \hline
    D$_2$: tupled+slotted \ & \ in{}5. D$_0$(m=1,k$\geq$1): shelframeships \ & \ in{}6. D$_0$(m=1,k$\geq$1): shelframepoints \ \! \\ \hline
  \end{tabular}
%
\\[0.3cm]
  \begin{tabular}{ l | l | l |}
    \textbf{D$_3$: dependent \ \ \ } & \ D$_1$: oidless \ & \ D$_1$: oidful \ \\ \hline
    D$_2$: tupled & \ de{}1. D$_0$(m=1,k=0): \textbf{\textit{relationships}} \! & \ de{}2. D$_0$(m=1,k=0): relationpoints \ \ \ \\ \hline
    D$_2$: slotted & \ de{}3: pairships & \ de{}4: pairpoints \\ \hline
    D$_2$: tupled+slotted \ & \ de{}5. D$_0$(m=1,k$\geq$1): relpairships & \ de{}6. D$_0$(m=1,k$\geq$1): relpairpoints \\ \hline
  \end{tabular}
%
\\[0.3cm]
  \begin{tabular}{ l | l | l |}
    \textbf{D$_3$: independent+dependent \ \ \ \ } & \ D$_1$: oidless \ \ \ \ \ \ \ \ \ \ \! & \ D$_1$: oidful \ \ \ \ \ \ \ \ \ \ \ \ \ \ \ \ \ \ \ \ \  \ \ \ \ \ \ \ \ \ \ \ \ \ \! \\ \hline
    D$_2$: tupled & \ id{}1 & \ id{}2 \\ \hline
    D$_2$: slotted & \ id{}3 & \ id{}4 \\ \hline
    D$_2$: tupled+slotted \ & \ id{}5 & \ id{}6 \\ \hline
  \end{tabular}
\end{small}
\caption{Basic metamodel of PSOA RuleML: Multi-dimensional psoa atoms.}
\label{Metamodel}
\end{figure*}

}
\end{appendix}

\end{document}